%% file: root.tex
\documentclass[letterpaper, 10 pt, conference]{ieeeconf}  

\IEEEoverridecommandlockouts                              

\usepackage{graphics} 
\usepackage{epsfig} 
\usepackage{mathptmx} 
\usepackage{times} 
\usepackage{amsmath} 
\usepackage{amssymb}  
\usepackage{url}
\usepackage{booktabs}
\usepackage{siunitx}
\usepackage{csquotes}
\usepackage{pgf}
\usepackage{pgfplots}
\usepackage[T1]{fontenc}

\usepackage{multirow} 
\usepackage{pifont} 
\usepackage{arydshln} 
\usepackage{subfiles}
\usepackage{balance} 
\usepackage[absolute]{textpos} 
\usepackage{xcolor}

\newcolumntype{x}[1]{>{\centering\arraybackslash\hspace{0pt}}p{#1}}

\usepackage{etoolbox}
\makeatletter
\patchcmd{\@makecaption}
  {\scshape}
  {}
  {}
  {}
\makeatletter
\patchcmd{\@makecaption}
  {\\}
  {.\ }
  {}
  {}
\let\NAT@parse\undefined
\makeatother

\title{\LARGE \bf
The GOOSE Dataset for Perception in Unstructured Environments
}

\usepackage[backref=page]{hyperref}
\hypersetup{colorlinks=true, 
	unicode=true, 
	linkcolor=[rgb]{0.10,0.05,0.67}, 
	citecolor=green, 
	filecolor=[rgb]{0.10,0.05,0.67}, 
	urlcolor=[RGB]{237, 15, 152} 
}

\let\oldtwocolumn\twocolumn
\renewcommand\twocolumn[1][]{%
    \oldtwocolumn[{#1}{
    \begin{center}
           \includegraphics[width=\linewidth]{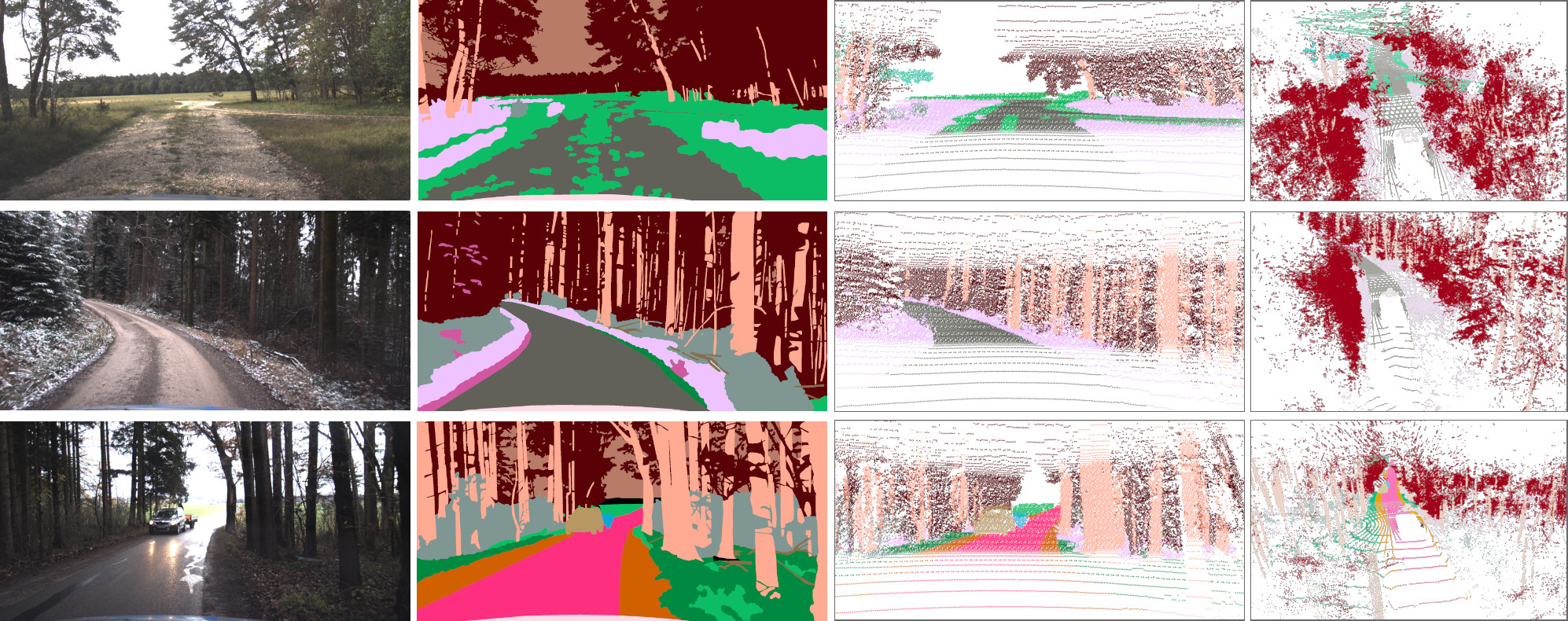}
           \scriptsize Fig. 1.~~The GOOSE dataset was recorded over the course of a year and covers all seasons and a wide range of weather conditions. The first column shows color images recorded with the RGB+NIR (near-infrared) camera. The second column is the respective pixel-wise annotated image. The third column shows the annotated point cloud from the 128-channel roof LiDAR system from the same perspective as the camera. The fourth column shows the same point cloud from a perspective outside of the survey vehicle.
        \end{center}
    }]
}

\newcommand{\copyrightstatement}{
	\begin{textblock*}{17cm}(20mm,1mm)    
		\noindent
		\footnotesize
		\copyright 2024 IEEE. Personal use of this material is permitted. Permission from IEEE must be
		obtained for all other uses, in any current or future media, including
		reprinting/republishing this material for advertising or promotional purposes, creating new
		collective works, for resale or redistribution to servers or lists, or reuse of any copyrighted
		component of this work in other works.
	\end{textblock*}
}

\newcommand{\underReviewStatement}{
	\begin{textblock*}{17cm}(20mm,1mm)    
		\noindent
		\footnotesize
		This is a preprint of a paper submitted to an IEEE journal/conference. 
     The paper is currently under review. 
     Please cite the final published version when available.
	\end{textblock*}
}

\newif\ifaddUnderReviewNotice
\addUnderReviewNoticefalse

\newif\ifaddIEEECopyrightNotice
\addIEEECopyrightNoticetrue

\author{Peter Mortimer$^{1}$, Raphael Hagmanns$^{2,3}$, Miguel Granero$^{2}$, \\ Thorsten Luettel$^{1}$, Janko Petereit$^{2}$ and Hans-Joachim Wuensche$^{1}$
\thanks{This work was supported by the Federal Office of Bundeswehr Equipment, Information Technology and In-Service Support (BAAINBw).}%
\thanks{$^{1}$ The authors are with the Institute for Autonomous Systems Technology, University of the Bundeswehr Munich, Germany
        {\tt\small firstname.lastname@unibw.de}}%
\thanks{$^{2}$ The authors are with the Fraunhofer Institute of Optronics, System Technologies and Image Exploitation, Karlsruhe, Germany {\tt\small firstname.lastname@iosb.fraunhofer.de}}%
\thanks{$^{3}$ The author is with the Karlsruhe Institute of Technology (KIT), Germany
        {\tt\small firstname.lastname@kit.edu}}%
}

\begin{document}

\ifaddUnderReviewNotice
\underReviewStatement
\fi

\ifaddIEEECopyrightNotice
\copyrightstatement
\fi

\setcounter{figure}{1}

\input{tex/sensor_colors}

\maketitle
\thispagestyle{empty}
\pagestyle{empty}

\begin{abstract}
The potential for deploying autonomous systems can be significantly increased by improving the perception and interpretation of the environment. 
However, the development of deep learning-based techniques for autonomous systems in unstructured outdoor environments poses challenges due to limited data availability for training and testing. 
To address this gap, we present the German Outdoor and Offroad Dataset (GOOSE), a comprehensive dataset specifically designed for unstructured outdoor environments. 
The GOOSE dataset incorporates 10\,000 labeled pairs of images and point clouds, which are utilized to train a range of state-of-the-art segmentation models on both image and point cloud data. 
We open source the dataset, along with an ontology for unstructured terrain, as well as dataset standards and guidelines. 
This initiative aims to establish a common framework, enabling the seamless inclusion of existing datasets and a fast way to enhance the perception capabilities of various robots operating in unstructured environments. 
This framework also makes it possible to query data for specific weather conditions or sensor setups from a database in future. 
The dataset, pre-trained models for offroad perception, and additional documentation can be found at \url{https://goose-dataset.de/}. 
\end{abstract}

\section{Introduction}
\label{sec:introduction}
Perception in unstructured outdoor environments poses significant challenges for autonomous robots. 
In offroad environments, mobile robots encounter complex decision-making scenarios that require a fine-grained understanding of their surroundings. 
One crucial aspect for safe and efficient navigation is the accurate categorization of free space as safely traversable, which requires distinguishing between various obstacle types. 
Depending on the robotic platform, some regions filled with high grass or bushes might be traversable while this might not be the case for other platforms. 
In addition, environments can be very different depending on weather conditions or other outer influences.
Therefore, most navigation solutions perform poorly in those environments. Many of the current solutions only provide semi-autonomous navigation, as they often rely on teach-and-repeat or similar strategies. 

We propose the GOOSE dataset as a novel dataset focused on the perception in unstructured outdoor environments. The size of the GOOSE dataset sets a new standard in annotated 3D point clouds and camera images for unstructured outdoor environments. This will allow the dataset to be a reliable measure of progress by quantitatively evaluating and comparing new methods while simultaneously forming the basis for learning-based methods from other research groups that can use the GOOSE dataset as source for transfer learning on their own robotics platforms.

\begin{table*}[htpb!]
\vspace{1em}
\centering
\caption{
Comparison of sizes and sensor modalities between existing offroad datasets. 
In terms of size, the RELLIS-3D provides more annotated laser scans, but fewer annotated images.
The individual scans in the GOOSE dataset are denser and were recorded with a 128-channel LiDAR scanner, which is a novelty among driving datasets.
}
\begin{tabular}{@{}ccccccl@{}}
\toprule
Dataset & Platform & Sensors & Annotated Sensor Modalities & \# Annotations & \# Classes \\ \midrule
OFFSED \cite{neigel_offsed_2021} & ZED & stereo camera & RGB+Depth & 203 & 19 \\
Freiburg Forest \cite{valada_deepscene_2016} & Viona & multi-spectral camera, stereo camera & RGB+NIR+Depth &     366 & 7 \\
TAS500 \cite{metzger_tas500_2020} & MuCAR-3 & camera & RGB & 640 & 24 \\
YCOR \cite{maturana_ycor_2018}  & ATV & camera & RGB & 1\,076 & 8 \\
RUGD \cite{wigness_rugd_2019} & Husky & camera  & RGB & 7\,546 & 24 \\
RELLIS-3D \cite{jiang_rellis-3d_2021} & Warthog & stereo camera / LiDAR / INS & RGB+Depth / Point Cloud & 6\,235 / 13\,556 & 20 \\
GOOSE (\textbf{ours}) & MuCAR-3 & prism camera / LiDAR / INS & RGB+NIR / Point Cloud & 10\,000 / 10\,000 & 64 \\ \bottomrule
\end{tabular}
\label{tab:datasets}
\end{table*}
\vspace{1em}
This paper presents several key contributions aimed at advancing the perception for unstructured environments:
\begin{itemize}
    \item We introduce the GOOSE dataset, which comprises of 10\,000 pixel-wise annotated pairs  of RGB images and LiDAR point clouds, offering a diverse and representative collection of data. The GOOSE dataset encompasses a wide range of environments and includes occurrences of 64 classes that are commonly encountered in unstructured outdoor settings.
    \item We provide open-source access to the data, including ROS bags and accompanying tools, making it readily available for robotic research. For that reason, additional sensor data, such as near-infrared (NIR) channels of the cameras, surround views, and a high-precision localization based on RTK-GNSS and IMU measurements, are provided to extend the research scope beyond semantic segmentation in 2D and 3D.
    \item We evaluate the performance of different state-of-the-art models for both 2D semantic segmentation on the RGB camera images and 3D semantic segmentation on the 3D point clouds from the LiDAR scanner.
\end{itemize}

\section{Related Work}
\label{sec:related-work}

Several datasets have been published for urban environments~\cite{geiger_vision_2013,cordts_cityscapes_2016,caesar_nuscenes_2020,behley_semantickitti_2019,zendel_wilddash2_2022}. Modern object detection and segmentation models have had great success in using these large datasets to improve perception capabilities for the classical autonomous driving application case. Typical sensor modalities of these datasets are RGB camera images or LiDAR point clouds, depending on the concrete application.
However, there have been very few attempts to replicate the results in urban environments for navigation in unstructured outdoor environments~\cite{neigel_offsed_2021,valada_deepscene_2016, metzger_tas500_2020, maturana_ycor_2018, wigness_rugd_2019, jiang_rellis-3d_2021}. 
There are only a few comparatively small datasets with limited sensor modalities, and there is no established ontology for relevant objects in such environments.
Table~\ref{tab:datasets} gives an overview of publicly available datasets and their main characteristics.

Not considered in the comparison in Table~\ref{tab:datasets} are datasets that were annotated in unstructured environments for other specific tasks such as offroad freespace detection~\cite{min_orfd_2022}, place recognition~\cite{knights_wildplaces_2023}, learning offroad dynamic models~\cite{triest_tartandrive_2022} or end-to-end driving~\cite{tampuu_estoniadriving_2023}.

Currently, the RUGD dataset~\cite{wigness_rugd_2019} includes the largest number of annotations primarily focused on offroad scenes. 
It has established an ontology that encompasses a variety of typical offroad environments, but the dataset exclusively consists of RGB images. 
In contrast, datasets like OFFSED~\cite{neigel_offsed_2021} and Freiburg Forest~\cite{valada_deepscene_2016} include additional multi-spectral and stereo data, although both have a limited number of annotated frames. 

Similar limitations apply to the TAS500 dataset~\cite{metzger_tas500_2020}, which was predominantly created for evaluation purposes but shares the same MuCAR-3 (Munich Cognitive Autonomous Robot Car) platform as GOOSE. 
On the other hand, the YCOR dataset~\cite{maturana_ycor_2018} offers scenes from four distinct locations, showcasing diverse lighting and weather conditions. 

In this context, the RELLIS-3D dataset~\cite{jiang_rellis-3d_2021} emerges as the largest contribution, incorporating multiple annotated modalities in a substantial number of frames, suitable for modern deep-learning pipelines. 
By providing both images and point clouds across a diverse range of scenes, it facilitates the training of advanced semantic segmentation models. 
Nevertheless, the RELLIS-3D dataset has a number of limitations. 
First, its ontology doesn't allow for detailed classification of trafficability. 
For instance, the label \textit{grass} constitutes a significant portion of the dataset in terms of pixels count, yet displays considerable visual variability. 
Furthermore, point clouds are labeled based solely on RGB camera annotations. Multiple views are fused using a SLAM approach for a higher labeling coverage, but the amount of labeled points is still limited. 
A RELLIS-3D scan comprises up to 13\,056 points, which is not much considering the sensor setup. 
Lastly, the RELLIS-3D dataset only encompasses scenes from a single environment, resulting in relatively consistent lighting conditions and general class appearances across settings. 
This is also reflected by the comparatively low number of 20 classes in the RELLIS-3D dataset, considering the high amount of annotations.

With the GOOSE dataset, we have tried to overcome these drawbacks. Thus, our point clouds are pointwise annotated and consist of more than 200\,000 points on average. We have also put a focus on diversity in scenes and environments, which is supported by our large 64 class ontology. Furthermore, both image and point clouds are annotated with corresponding instances, which enables a combination of 2D and 3D based perception. We therefore expect that the GOOSE dataset can contribute to deep-learning based navigation in a variety of offroad environments.

\section{The GOOSE Dataset}
\label{sec:goose-dataset}

\subsection{Ontology}

The GOOSE dataset uses 64 semantic classes in total to segment the RGB images and LiDAR point clouds. The ontology presented in Table~\ref{tab:class_overview} is inspired by similar autonomous driving datasets such as SemanticKITTI \cite{behley_semantickitti_2019}, TAS500 \cite{metzger_tas500_2020} and RELLIS-3D \cite{jiang_rellis-3d_2021}.
The semantic segmentation of the GOOSE dataset is generally more granular, in particular when segmenting the terrain surface and vegetation types.
Many semantic classes can be grouped into broader parent semantic classes, such as $\{$\textit{moss}, \textit{low grass}, \textit{leaves}, \textit{high grass}$\}$ $\rightarrow$ \textit{drivable vegetation}.
This composability is motivated by the instance labels of the ATLAS ontology \cite{smith_atlas_2022}, an all-terrain label set for autonomous systems. 
The fine-granular segmentation also makes the GOOSE dataset compatible for composite datasets for multi-domain semantic segmentation like MSeg \cite{lambert_mseg_2020}.


\input{tex/goose_classes_table.tex}


\subsection{Dataset Statistics}
\label{subsec:dataset-statistics}
\begin{figure*}[htpb]
  \centering
  \resizebox{\textwidth}{!}{\input{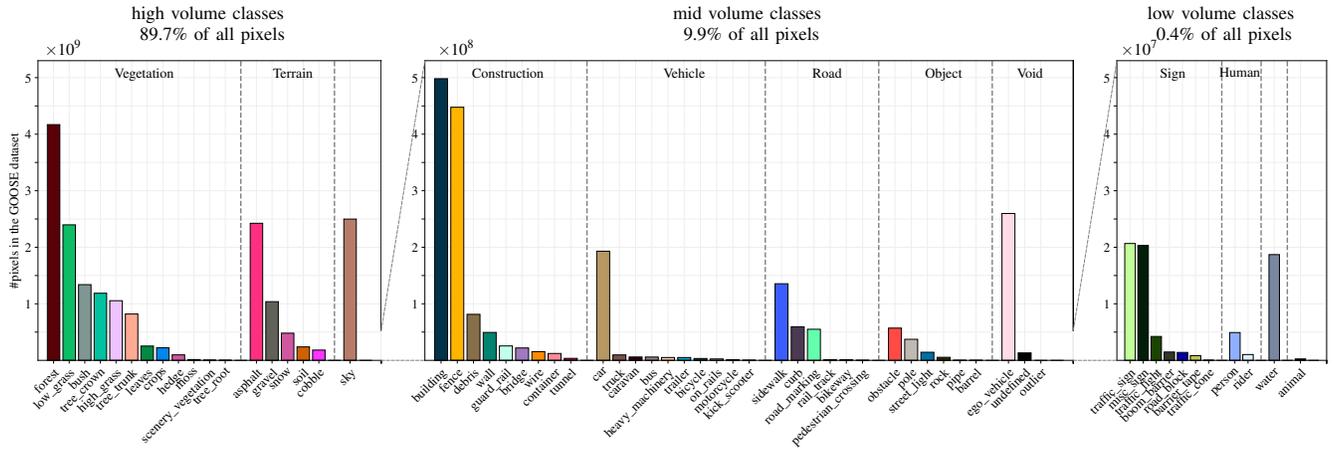}}

  \caption{Best to inspect digitally. Histogram of the annotated pixels in the 2D images of the GOOSE dataset. The 12 categories can be grouped into the high volume classes in the categories $\{$\textit{Vegetation}, \textit{Terrain}, \textit{Sky}$\}$, mid volume classes from the categories $\{$\textit{Construction}, \textit{Vehicle}, \textit{Road}, \textit{Object}, \textit{Void}$\}$ and the low volume classes from the categories $\{$\textit{Sign}, \textit{Human}, \textit{Water}, \textit{Animal}$\}$. }
  \label{fig:2d_histogram}
\end{figure*}

\begin{figure*}[htpb]
  \centering
  \resizebox{\textwidth}{!}{\input{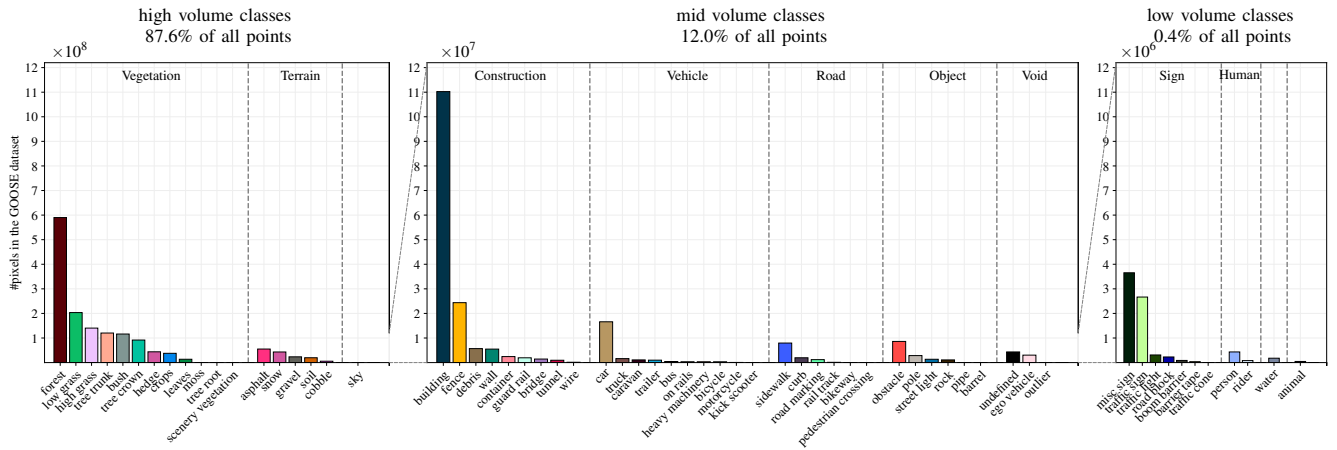}}
  \caption{
  Best to inspect digitally. 
  Histogram of the annotated points in the 3D point clouds of the GOOSE dataset. 
  We observe a relative order of the semantic classes similar to that of the annotated images in the GOOSE data set (see Figure~\ref{fig:2d_histogram}). The elevated position of the 128-channel LiDAR scanner leads to many detections of building roofs, which explains the high occurrence of annotated \textit{building} points among the mid volume classes.
  }
  \label{fig:3d_histogram}
\end{figure*}

The GOOSE dataset consists of pixel-annotated images of a RGB+NIR front-facing camera and an annotated point cloud from the spinning LiDAR system with 128 channels on the roof of MuCAR-3 and two side-facing LiDAR systems with 32 channels. 
Figure~\ref{fig:2d_histogram} shows a histogram of all annotated pixels in the image data of the GOOSE dataset.
Almost 90\% of the annotated pixels are in the categories \textit{vegetation}, \textit{terrain} and \textit{sky}. This is consistent with our goal of recording mainly in unstructured outdoor environments. 
The remaining 10\% of the annotated pixels make up different obstacle types like \textit{fence}, \textit{building} and \textit{pole}, as well as dynamic objects like \textit{car}, \textit{person} and \textit{animal}.
Figure~\ref{fig:3d_histogram} shows the semantic histogram of the annotated points in the 3D point cloud, which has a very similar relative ordering to the annotated pixels.
The GOOSE dataset is very fine-granular with its segmentation into 64 semantic classes.
One caveat caused by the fine-granular segmentation in the GOOSE dataset, is that some semantic classes only rarely appear in the dataset like \textit{water} or \textit{rider}.

\subsection{Annotation Process}
\label{subsec:annotation-process}
The RGB image of the RGB+NIR camera was annotated by a human annotator. 
The annotation consistency was ensured by providing a detailed labeling policy beforehand, which will be co-released on the GOOSE dataset website.

Multiple LiDAR scans from the same sequence are fused in the same coordinate frame.
This allows the human annotator to more easily recognize objects in the far distance with a higher point density.
The instance IDs of the objects in the annotated RGB image and LiDAR scans of the same timestamp are then matched.

\subsection{Data Format}
\label{subsec:data-format}
We provide all raw sensor data from each sequence in the ROS bag format~\cite{quigley_ros1_2009}.
The annotated sensor data in a given sequence is published as a ROS message.
Additionally, the annotated image and point cloud data is available in a format similar to SemanticKITTI~\cite{behley_semantickitti_2019}.


\section{Robot Setup}
\label{sec:robot-setup}

\subsection{Sensor Setup}
\label{subsec:sensor-setup}
\begin{figure}[htpb!]
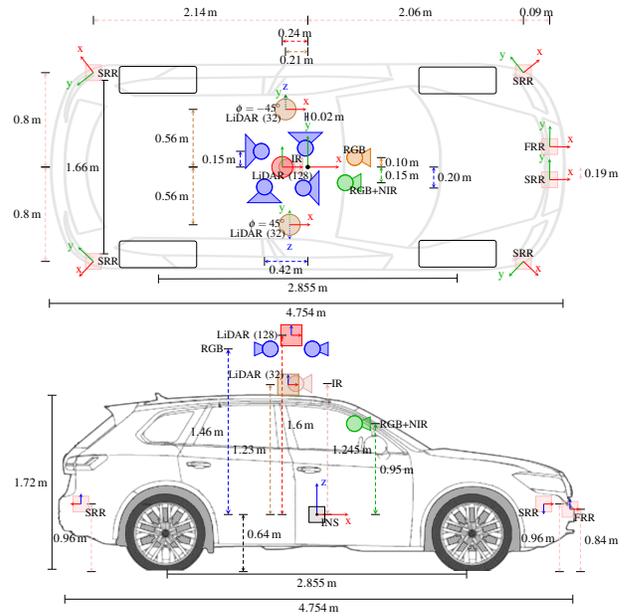

  \centering
  \subfile{images/sensor_setup/birds_eye_view}
  \subfile{images/sensor_setup/side_view}
  \caption{
  Schematic of the sensor setup on MuCAR-3. MuCAR-3 is a modified Volkswagen Touareg with full drive-by-wire capabilities. 
  The base\_link coordinate frame is situated between both axles at the vehicle's center of gravity and coincides with the \textbf{INS}~(\textcolor{ins}{$\blacksquare$}) position. 
  The \textbf{RGB+NIR}~(\textcolor{windshield}{$\blacksquare$}) camera image is annotated as part of the GOOSE dataset and is mounted behind the windshield. 
  The point cloud of the \textbf{128-beam LiDAR}~(\textcolor{vls128}{$\blacksquare$}) scanner on the roof of MuCAR-3 is also annotated.  
  The two \textbf{32-beam LiDAR}~(\textcolor{os32}{$\blacksquare$}) scanners are tilted at an $\pm\SI{45}{\degree}$ angle to detect close by obstacles lateral to MuCAR-3. These points are also annotated for the same timestamp in the GOOSE dataset.
  The four \textbf{RGB}~(\textcolor{surround}{$\blacksquare$}) color cameras on the roof provide a \SI{360}{\degree} view of the surrounding. 
  A front-facing \textbf{thermal infrared (IR)}~(\textcolor{ir}{$\blacksquare$}) camera provides an additional modality for low-light conditions.
  Six \textbf{radar}~(\textcolor{radar}{$\blacksquare$}) sensors are mounted around the vehicle to provide 360 degree radar detections with only small blind spots on the sides.
  An additional \textbf{RGB}~(\textcolor{marveye}{$\blacksquare$}) color camera mounted on a mobile platform~\cite{unterholzner_marveye_2010} is positioned behind the windshield and pans the camera when turning into corners or following objects of interest.
  }
  \label{fig:sensor_setup}
\end{figure}

The GOOSE dataset was recorded on the UniBw Munich research vehicle MuCAR-3~\cite{himmelsbach_mucar3_2011}. MuCAR-3 is a modified Volkswagen Touareg with full drive-by-wire capabilities. The sensor setup of MuCAR-3 is illustrated in Figure~\ref{fig:sensor_setup} and includes:
\begin{itemize}
    \item 1 $\times$ Velodyne Alpha Prime (\textcolor{vls128}{$\blacksquare$}): 128 channels, 10~\si{\hertz}, \SI{40}{\degree} vertical field of view
    \item 2 $\times$ Ouster OS0 (\textcolor{os32}{$\blacksquare$}): 32 channels, \SI{10}{\hertz}, \SI{90}{\degree} vertical field of view
    \item 4 $\times$ Basler acA2440-20gc (\textcolor{surround}{$\blacksquare$}): RGB camera with a \SI{4}{\milli\meter} 1/1.8\textquotedbl~Lensation BM4018S118C lens, \SI{10}{\hertz}, \SI{94}{\degree} horizontal field of view 
    \item 1 $\times$ FLIR A615 (\textcolor{ir}{$\blacksquare$}): thermal infrared camera with a built-in lens, \SI{50}{\hertz}, \SI{45}{\degree} horizontal field of view
    \item 1 $\times$ JAI FSFE-3200D (\textcolor{windshield}{$\blacksquare$}): prism camera with a sensor behind a RGB color filter and a sensor behind a near infrared filter with a \SI{6}{\milli\meter} 1/1.8\textquotedbl~Fujinon TF6MA-1 lens, \SI{10}{\hertz}, \SI{59}{\degree} horizontal field of view
    \item 1 $\times$ Basler acA2440-20gc (\textcolor{marveye}{$\blacksquare$}): RGB camera with a \SI{8}{\milli\meter} 1/1.4\textquotedbl~Kowa LM8HC mounted on a mobile platform, \SI{10}{\hertz}, \SI{54}{\degree} horizontal field of view  
    \item 5 $\times$ Smartmicro UMRR-96 Type 153 (\textcolor{radar}{$\blacksquare$}): \SI{79}{\giga\hertz} band, set in medium-range mode with a detection range of \SI{0.4}{\meter}-\SI{55}{\meter}
    \item 1 $\times$ Smartmicro UMRR-11 Type 132 (\textcolor{radar}{$\blacksquare$}): \SI{77}{\giga\hertz} band, long-range mode with a detection range of \SI{1}{\meter}-\SI{175}{\meter}
    \item 1 $\times$ Oxford RT3000v3 (\textcolor{ins}{$\blacksquare$}): Inertial Navigation System (INS) with differential RTK-GNSS corrections received over LTE using the NTRIP protocol
\end{itemize}
The low level vehicle control and mobile camera platform control is handled by two real-time capable dSPACE computers. The main computing unit for sensor data accumulation and recording is equipped with a Titan RTX GPU and an AMD Ryzen Threadripper PRO 3975WX CPU.

\subsection{Synchronization}
\label{subsec:synchronization}

The sensor clocks of the cameras, LiDAR scanners and INS system are synchronized using the Precision Time Protocol (PTP, IEEE 1588~\cite{ieee_ptp}). In essence, PTP synchronizes clocks across a sensor network by determining a leader clock in a local network and synchronizing the follower clocks to the leader clock while also taking the message delay into consideration.

We predetermined our INS system as leader clock. The INS obtains its time at \SI{100}{\hertz} over GNSS. The color cameras from Basler and JAI, the main computer's system clock, as well as the side-facing Ouster OS0 LiDAR sensors support the synchronization of their clocks using PTP. This synchronization also allows for triggering the cameras with respect to the current LiDAR orientation


\subsection{Calibration}
\label{subsec:calibration}

The intrinsic parameters of each camera are determined by calibrating them on a mosaic of AprilTags \cite{wang_apriltag2_2016} with checkerboard corners as described in~\cite{kallwies_apriltagcorners_2020}. All cameras are assumed to fit a pinhole projection model. 
The extrinsic calibration of the sensors is essential for sensor fusion. 
The relative transformation between the roof LiDAR scanner and color cameras is determined using a calibration target with four circular holes, which can be automatically detected in the LiDAR point cloud~\cite{guindel_lidarcalibrationboard_2017}. 
Similar to~\cite{beltran_lidarcameracalibration_2022}, the calibration target has four additional AprilTags, which can be detected accurately in the color camera images. Detailed calibration procedures will be made available in the dataset documentation.

\vspace{2em}
\section{Experimental Evaluation}
\label{sec:evaluation}

For the initial evaluations, we decided to focus on the 2D image segmentation from the RGB camera image (see Subsection~\ref{subsec:image-segmentation}) and 3D point cloud segmentation from a single LiDAR scan (see Subsection~\ref{subsec:point-cloud-segmentation}).

\subsection{Training Split}
\label{subsec:training-split}

The GOOSE dataset was recorded over the course of a year.
Each recording day is considered a \textit{scenario} consisting of multiple \textit{sequences}.
The raw sensor data of a sequence is stored in ROS bag format for later reprocessing.
To use the GOOSE dataset as a benchmark, we define a fixed training, validation and test split for supervised learning algorithms.
We ensured that each split contains sequences from each season and a wide range of weather conditions, while still sharing a similar distribution among the semantic classes that appear in each split. 
The training split holds 7830 annotated images and point clouds, the validation split 960 and the test split 1210.
The annotations of the test split are withheld from the publication of the GOOSE dataset to allow the setup of a public benchmarking platform in the future.

\begin{table*}[htpb!]
\vspace{1em}
\centering
    \caption{Comparison of the 2D image segmentation and 3D point cloud segmentation performance on the GOOSE test set. The IoU scores are specified in percent. For class-based evaluation, classes with occurences less than 20 are omitted. No classes of the category \textit{Sky} exist for the 3D point cloud segmentation.}
    \renewcommand{\arraystretch}{1.2}
    \begin{tabular}{*{8}{cllccccccccccc}}
        \toprule
        & network & type & \textbf{mIoU} & Vegetation & Terrain & Vehicle & Object & Constr. & Road & Sign & Human & Sky & Water \\
        
        \midrule
        \multirow{5}{*}{\rotatebox[origin=c]{90}{2D}} & 
        \multirow{2}{*}{PP-LiteSeg \cite{peng_ppliteseg_2022}}
        & \textit{category}
        & \textbf{67.21}
        & 92.43 
        & 90.62 
        & 87.83 
        & 46.86 
        & 78.81 
        & 51.23 
        & 58.19 
        & 60.53 
        & 96.28 
        & 9.29 
        \\
        &
        & \textit{class}
        & \textbf{45.09}
        &  47.69 
        &  70.97 
        &  37.69 
        &  29.15 
        &  43.96 
        &  49.71 
        &  33.48 
        &  34.07 
        &  96.11 
        &  8.23 
        \\
        & \multirow{2}{*}{DDRNet~\cite{pan_ddrnet_2022}}
        & \textit{category}
        & \textbf{70.23}
        & 92.94 
        & 91.34 
        & 91.05 
        & 52.66 
        & 81.67 
        & 59.27 
        & 62.90 
        & 64.65 
        & 96.55 
        & 9.23 
        \\
        &
        & \textit{class}
        & \textbf{46.53}
        & 49.54 
        & 72.20 
        & 31.22 
        & 35.63 
        & 44.52 
        & 54.50 
        & 37.97 
        & 34.38 
        & 96.24 
        & 9.09 
        \\
        & Mask2Former~\cite{cheng2021mask2former}
        & \textit{category}
        & \textbf{64.26}
        & 91.76 
        & 91.34 
        & 91.05 
        & 42.57 
        & 75.07 
        & 44.26 
        & 43.22 
        & 61.72 
        & 96.49 
        & 5.10 
        \\
        \midrule
        \multirow{2}{*}{\rotatebox[origin=c]{90}{3D}} & 
        PVKD \cite{hou_pointvoxelknowledgedistill_2022, zhu_cylinder3d_2021} 
        & \textit{class}
        & \textbf{34.32} 
        & 47.40 
        & 28.69 
        & 40.82 
        & 20.61 
        & 37.94 
        & 36.21 
        & 29.89 
        & 57.94 
        & - 
        & 9.42 
        \\ 
        & SPVNAS \cite{tang_spvnas_2020} 
        & \textit{class}
        & \textbf{17.32} 
        & 41.08 
        & 19.27 
        & 14.58 
        & 21.29 
        & 21.27 
        & \phantom{0}6.38 
        & 12.94 
        & 19.08 
        & - 
        & \phantom{0}0.0 
        \\ 
        \bottomrule
    \end{tabular}
    \label{tab:evaluation}
\end{table*}

\subsection{Evaluation Metrics}
\label{subsec:evaluation-metrics}

The standard metric for evaluating the semantic segmentation on both images and point clouds is the mean Intersection over Union (mIoU), also known as the mean Jaccard index. The IoU is widely used for evaluating the segmentation of a semantic class by comparing the prediction region with the ground truth region. It can be described using the result cases true positive (TP), false negative (FN) and false positive (FP) of a binary classification in the following way:
\begin{align}
    \text{IoU} = \frac{\text{TP}}{\text{TP} + \text{FN} + \text{FP}} \, .
\end{align}
The IoU is determined by accumulating TP, FP and FN of the whole test set. The mIoU is then calculated by averaging the IoU over all classes in the dataset.
The IoU is known to be biased towards object instances and classes that cover large areas of the image \cite{cordts_cityscapes_2016}. 
In Table~\ref{tab:evaluation}, we provide averaged IoU values for models which were trained on the full set of classes \textit{(class)} as well as IoU values for models trained on the broader category labels \textit{(category)}.

\subsection{Image Segmentation}
\label{subsec:image-segmentation}

Our selection of image semantic segmentation algorithms was determined by comparing the state-of-the-art methods on other public benchmarks.
Since the GOOSE dataset will primarily be used in a outdoor robotics context, inference speed is also an important criterion for method selection.
DDRNet~\cite{pan_ddrnet_2022} uses a typical two-stream architecture originated in BiSeNet~\cite{yu_bisenet_2018}, but fuses both branches at different depths in the network. 
PPLiteSeg~\cite{peng_ppliteseg_2022} uses an encoder-decoder structure with a lightweight attention-based fusion model in the decoder to enable realtime semantic segmentation.
Mask2former~\cite{cheng2021mask2former} is a recent, universal, transformer-based model that can be conditioned at both semantic and instance segmentation.

From the results in Table~\ref{tab:evaluation} we observe a solid image segmentation quality in general. 
Most categories with many appearances have IoU values over 90\% (\textit{Vegetation}, \textit{Terrain}, \textit{Vehicle}, \textit{Sky}).
However, we also encounter some categories with very small IoU values, leading to a reduced value (10- 20\%) for the overall mIoU compared to the same algorithm applied on the Cityscapes~\cite{cordts_cityscapes_2016} dataset.
The main reason for this is the natural imbalance in the class distribution within our dataset. 
Limited training data availability as well as similar appearances of different classes lead to difficulties in recognizing those classes correctly. The comparatively low class-based IoU values in Table~\ref{tab:evaluation} show that performing fine-graded segmentation is difficult in environments, where the transitions between regions are not always sharp.
These effects might be mitigated by applying typical class balancing strategies, such as weighted loss functions or specialized fine-graded architectures, which will be further explored in future research. 
Although classes with only a few labels may have a negative impact on the mIoU score, their existence is highly relevant for practical navigation solutions since misclassified regions such as \textit{water} may also lead to wrong classification of trafficability. 

\subsection{Point Cloud Segmentation}
\label{subsec:point-cloud-segmentation}

For the evaluation of the annotated 3D point cloud data, we focus on methods that produce an annotated point cloud instead of a voxel grid or a birds-eve-view segmentation of the surroundings.
Point-Voxel-KD~\cite{hou_pointvoxelknowledgedistill_2022} (PVKD) trains a slim student network by mimicking a larger teacher model both in its voxel and point representation. The student and teacher architectures are based on Cylinder3D~\cite{zhu_cylinder3d_2021}.
SPVNAS~\cite{tang_spvnas_2020} propose the Sparse Point-Voxel Convolution (SPVConv) block that makes use of both a point and a voxel-based representation, while also searching for an efficient network architecture of SPVConv blocks through neural architecture search (NAS).

In the lower part of Table~\ref{tab:evaluation}, we compare the 3D segmentation performance of Point-Voxel-KD and SPVNAS on the GOOSE dataset. As input, both models use the 3D position and the reflection intensity of each point in the point cloud.
For both models, we observe low mIoU scores for rare categories like \textit{Human}, \textit{Sky} and \textit{Water}. 
Categories like \textit{Terrain} contain many classes that are geometrically similar and differ mainly in their surface texture like \textit{gravel}, \textit{soil} and \textit{snow}.
In these categories, we also observe that both models perform poorly compared to the image segmentation results for the same category.

\section{Conclusion}
\label{sec:conclusion}

We introduce the GOOSE dataset, an open-source, multimodal dataset for perception in unstructured outdoor environments.
As our benchmarks show, GOOSE dataset provides challenges in terms of imbalanced class distributions across a diverse set of outdoor environments. 
In our future work, we want to investigate additional semantic segmentation models, particularly those that can utilize different modalities. 
Additionally, we will expand the GOOSE-dataset by adding data from additional platforms in different environments to further enhance the adaptability of segmentation techniques for mobile robot navigation.\\
\\
\noindent\textbf{Acknowledgements} 
We thank our UniBw Munich student research assistant Abhishek Kotha who contributed with the verification of the annotated data. 
We also thank our colleague Bianca Forkel for her AprilTag-based LiDAR-camera calibration software.
Finally, we also thank our collaborators Roman Abayev and Anselm von Gladiß from the Active Vision Group (AGAS) at the University of Koblenz for providing the GOOSE DB, a database to query for specific recordings within the GOOSE dataset. 

\balance
\bibliographystyle{bib/IEEEtran}
\bibliography{bib/IEEEfull,bib/additional_full,bib/literature}

\end{document}

%% file: tex/sensor_colors.tex
\definecolor{windshield}{RGB}{72, 199, 72}
\definecolor{vls128}{RGB}{252, 46, 46}
\definecolor{surround}{RGB}{29, 29, 255}
\definecolor{os32}{RGB}{197, 141, 83}
\definecolor{radar}{RGB}{255, 215, 215}
\definecolor{ir}{RGB}{233, 180, 180}
\definecolor{marveye}{RGB}{236, 146, 56}
\definecolor{ins}{RGB}{62, 62, 62}

%% file: tex/goose_classes_table.tex
\input{tex/goose_colors.tex}

\begin{table}
    \caption{An overview of all 64 classes in the GOOSE dataset. The semantic classes were selected to cover both outdoor and urban driving scenarios. The most common \emph{thing} classes in the dataset are also segmented by instance (\ding{51}), allowing the evaluation of panoptic segmentation tasks. The pixelwise and pointwise annotations gain additional context by the annotated \emph{stuff} classes~\cite{caesar_cocostuff_2018}.}
    \hspace{-.8em}
	\begin{tabular}{|x{1.2cm}|x{2.3cm}|x{.5cm}x{1.6cm}|x{1cm}|}
    \hline
		category & class & \multicolumn{2}{|c|}{color} &instance\\ \hline
		\multirow{12}{*}{vegetation} & bush & \textcolor{bush}{$\blacksquare$} & (128,150,147) & \ding{53} \\
		& crops & \textcolor{crops}{$\blacksquare$} & (0,134,237) & \ding{53} \\
		& forest & \textcolor{forest}{$\blacksquare$} & (90,0,7) & \ding{53} \\
		& hedge & \textcolor{hedge}{$\blacksquare$} & (232,211,23) & \ding{53} \\
		& high\_grass & \textcolor{high_grass}{$\blacksquare$} & (238,195,255) & \ding{53} \\
		& leaves & \textcolor{leaves}{$\blacksquare$} & (0,137,65) & \ding{53} \\
		& low\_grass & \textcolor{low_grass}{$\blacksquare$} & (12,189,102) & \ding{53} \\
		& moss & \textcolor{moss}{$\blacksquare$} & (180,168,189) & \ding{53} \\
		& scenery\_vegetation & \textcolor{scenery_vegetation}{$\blacksquare$} & (69,109,117) & \ding{53} \\
		& tree\_crown & \textcolor{tree_crown}{$\blacksquare$} & (0,194,160) & \ding{53} \\
		& tree\_root & \textcolor{tree_root}{$\blacksquare$} & (196,164,132) & \ding{53} \\
		& tree\_trunk & \textcolor{tree_trunk}{$\blacksquare$} & (255,170,146) & \ding{51} \\ 
		\hdashline
		\multirow{5}{*}{terrain} & asphalt & \textcolor{asphalt}{$\blacksquare$} & (255,47,128) & \ding{53} \\
		& cobble & \textcolor{cobble}{$\blacksquare$} & (255,52,255) & \ding{53} \\
		& gravel & \textcolor{gravel}{$\blacksquare$} & (97,97,90) & \ding{53} \\
		& snow & \textcolor{snow}{$\blacksquare$} & (209,87,160) & \ding{53} \\
		& soil & \textcolor{soil}{$\blacksquare$} & (209,97,0) & \ding{53} \\
		\hdashline
		sky & sky & \textcolor{sky}{$\blacksquare$} & (183,123,104) & \ding{53} \\ 
		\hdashline
		\multirow{9}{*}{construction} & bridge & \textcolor{bridge}{$\blacksquare$} & (160,121,191) & \ding{53} \\
		& building & \textcolor{building}{$\blacksquare$} & (1,51,73) & \ding{53} \\
		& container & \textcolor{container}{$\blacksquare$} & (255,138,154) & \ding{51} \\
		& debris & \textcolor{debris}{$\blacksquare$} & (136,111,76) & \ding{53} \\
		& fence & \textcolor{fence}{$\blacksquare$} & (255,181,0) & \ding{53} \\
		& guard\_rail & \textcolor{guard_rail}{$\blacksquare$} & (194,255,237) & \ding{53} \\
		& tunnel & \textcolor{tunnel}{$\blacksquare$} & (204,7,68) & \ding{53} \\
		& wall & \textcolor{wall}{$\blacksquare$} & (0,132,111) & \ding{53} \\
		& wire & \textcolor{wire}{$\blacksquare$} & (255,140,0) & \ding{53} \\
		\hdashline
		\multirow{10}{*}{vehicle} & bicycle & \textcolor{bicycle}{$\blacksquare$} & (0,77,67) & \ding{51} \\
		& bus & \textcolor{bus}{$\blacksquare$} & (153,125,135) & \ding{51} \\
		& car & \textcolor{car}{$\blacksquare$} & (183,151,98) & \ding{51} \\
		& caravan & \textcolor{caravan}{$\blacksquare$} & (48,0,24) & \ding{51} \\
		& heavy\_machinery & \textcolor{heavy_machinery}{$\blacksquare$} & (250,208,159) & \ding{51} \\
		& kick\_scooter & \textcolor{kick_scooter}{$\blacksquare$} & (111,0,98) & \ding{51} \\
		& motorcycle & \textcolor{motorcycle}{$\blacksquare$} & (79,198,1) & \ding{51} \\
		& on\_rails & \textcolor{on_rails}{$\blacksquare$} & (161,194,153) & \ding{51} \\
		& trailer & \textcolor{trailer}{$\blacksquare$} & (10,166,216) & \ding{51} \\
		& truck & \textcolor{truck}{$\blacksquare$} & (123,79,75) & \ding{51} \\
		\hdashline
		\multirow{6}{*}{road} & bikeway & \textcolor{bikeway}{$\blacksquare$} & (163,0,89) & \ding{53} \\
		& curb & \textcolor{curb}{$\blacksquare$} & (74,59,83) & \ding{53} \\
		& pedestrian\_crossing & \textcolor{pedestrian_crossing}{$\blacksquare$} & (122,73,0) & \ding{53} \\
		& rail\_track & \textcolor{rail_track}{$\blacksquare$} & (107,121,0) & \ding{53} \\
		& road\_marking & \textcolor{road_marking}{$\blacksquare$} & (99,255,172) & \ding{53} \\
		& sidewalk & \textcolor{sidewalk}{$\blacksquare$} & (59,93,255) & \ding{53} \\
		\hdashline
		\multirow{6}{*}{object} & barrel & \textcolor{barrel}{$\blacksquare$} & (208,208,0) & \ding{51} \\
		& obstacle & \textcolor{obstacle}{$\blacksquare$} & (255,74,70) & \ding{53} \\
		& pipe & \textcolor{pipe}{$\blacksquare$} & (221,0,0) & \ding{53} \\
		& pole & \textcolor{pole}{$\blacksquare$} & (192,185,178) & \ding{51} \\
		& rock & \textcolor{rock}{$\blacksquare$} & (55,33,1) & \ding{51} \\
		& street\_light & \textcolor{street_light}{$\blacksquare$} & (0,111,166) & \ding{51} \\
		\hdashline
		\multirow{3}{*}{void} & ego\_vehicle & \textcolor{ego_vehicle}{$\blacksquare$} & (255,219,229) & \ding{53} \\
		& outlier & \textcolor{outlier}{$\blacksquare$} & (120,141,102) & \ding{53} \\
		& undefined & \textcolor{undefined}{$\blacksquare$} & (0,0,0) & \ding{53} \\
		\hdashline
		\multirow{7}{*}{sign} & barrier\_tape & \textcolor{barrier_tape}{$\blacksquare$} & (190,196,89) & \ding{53} \\
		& boom\_barrier & \textcolor{boom_barrier}{$\blacksquare$} & (52,54,45) & \ding{51} \\
		& misc\_sign & \textcolor{misc_sign}{$\blacksquare$} & (0,30,9) & \ding{51} \\
		& road\_block & \textcolor{road_block}{$\blacksquare$} & (0,0,166) & \ding{51} \\
		& traffic\_cone & \textcolor{traffic_cone}{$\blacksquare$} & (255,255,0) & \ding{53} \\
		& traffic\_light & \textcolor{traffic_light}{$\blacksquare$} & (27,68,0) & \ding{51} \\
		& traffic\_sign & \textcolor{traffic_sign}{$\blacksquare$} & (194,255,153) & \ding{51} \\
		\hdashline
		\multirow{2}{*}{human} & person & \textcolor{person}{$\blacksquare$} & (143,176,255) & \ding{51} \\
		& rider & \textcolor{rider}{$\blacksquare$} & (221,239,255) & \ding{51} \\
		\hdashline
		water & water & \textcolor{water}{$\blacksquare$} & (122,135,161) & \ding{53} \\
		\hdashline
		animal & animal & \textcolor{animal}{$\blacksquare$} & (0,0,53) & \ding{51} \\
		\hline
	\end{tabular}
 \label{tab:class_overview}
\end{table}

%% file: tex/goose_colors.tex
\definecolor{undefined}{RGB}{0,0,0}
\definecolor{traffic_cone}{RGB}{255,255,0}
\definecolor{snow}{RGB}{209,87,160}
\definecolor{cobble}{RGB}{255,52,255}
\definecolor{obstacle}{RGB}{255,74,70}
\definecolor{leaves}{RGB}{0,137,65}
\definecolor{street_light}{RGB}{0,111,166}
\definecolor{bikeway}{RGB}{163,0,89}
\definecolor{ego_vehicle}{RGB}{255,219,229}
\definecolor{pedestrian_crossing}{RGB}{122,73,0}
\definecolor{road_block}{RGB}{0,0,166}
\definecolor{road_marking}{RGB}{99,255,172}
\definecolor{car}{RGB}{183,151,98}
\definecolor{bicycle}{RGB}{0,77,67}
\definecolor{person}{RGB}{143,176,255}
\definecolor{bus}{RGB}{153,125,135}
\definecolor{forest}{RGB}{90,0,7}
\definecolor{bush}{RGB}{128,150,147}
\definecolor{traffic_light}{RGB}{27,68,0}
\definecolor{motorcycle}{RGB}{79,198,1}
\definecolor{sidewalk}{RGB}{59,93,255}
\definecolor{curb}{RGB}{74,59,83}
\definecolor{asphalt}{RGB}{255,47,128}
\definecolor{gravel}{RGB}{97,97,90}
\definecolor{boom_barrier}{RGB}{52,54,45}
\definecolor{rail_track}{RGB}{107,121,0}
\definecolor{tree_crown}{RGB}{0,194,160}
\definecolor{tree_root}{RGB}{196,164,132}
\definecolor{tree_trunk}{RGB}{255,170,146}
\definecolor{debris}{RGB}{136,111,76}
\definecolor{crops}{RGB}{0,134,237}
\definecolor{soil}{RGB}{209,97,0}
\definecolor{rider}{RGB}{221,239,255}
\definecolor{animal}{RGB}{0,0,53}
\definecolor{truck}{RGB}{123,79,75}
\definecolor{on_rails}{RGB}{161,194,153}
\definecolor{caravan}{RGB}{48,0,24}
\definecolor{trailer}{RGB}{10,166,216}
\definecolor{building}{RGB}{1,51,73}
\definecolor{wall}{RGB}{0,132,111}
\definecolor{rock}{RGB}{55,33,1}
\definecolor{fence}{RGB}{255,181,0}
\definecolor{guard_rail}{RGB}{194,255,237}
\definecolor{bridge}{RGB}{160,121,191}
\definecolor{tunnel}{RGB}{204,7,68}
\definecolor{pole}{RGB}{192,185,178}
\definecolor{traffic_sign}{RGB}{194,255,153}
\definecolor{misc_sign}{RGB}{0,30,9}
\definecolor{barrier_tape}{RGB}{190,196,89}
\definecolor{kick_scooter}{RGB}{111,0,98}
\definecolor{low_grass}{RGB}{12,189,102}
\definecolor{high_grass}{RGB}{238,195,255}
\definecolor{scenery_vegetation}{RGB}{69,109,117}
\definecolor{sky}{RGB}{183,123,104}
\definecolor{water}{RGB}{122,135,161}
\definecolor{wire}{RGB}{255,140,0}
\definecolor{outlier}{RGB}{120,141,102}
\definecolor{heavy_machinery}{RGB}{250,208,159}
\definecolor{container}{RGB}{255,138,154}
\definecolor{hedge}{RGB}{232, 211, 23}
\definecolor{moss}{RGB}{180, 168, 189}
\definecolor{barrel}{RGB}{208, 208, 0}
\definecolor{pipe}{RGB}{221, 0, 0}
\definecolor{military_vehicle}{RGB}{64,64,64}

%% file: images/sensor_setup/birds_eye_view.tex
\def\mucarDreiReifendurchmesser{.737}
\def\mucarDreiReifenbreite{.255}
\def\mucarDreiWidth{1.928}
\def\mucarDreiLengthRefToFrontEnd{2.3535}
\def\mucarDreiLengthRefToRearEnd{-2.4005}
\def\mucarDreiLengthRefToFrontAxle{1.4275}
\def\mucarDreiLengthRefToRearAxle{-1.4275}
\def\mucarDreiWidthRefToLeftFrontWheel{.8265}
\def\mucarDreiWidthRefToRightFrontWheel{-.8265}
\def\mucarDreiWidthRefToLeftRearWheel{.83225}
\def\mucarDreiWidthRefToRightRearWheel{-.83225}
\def\mucarThreeWheelbase{-\mucarDreiLengthRefToRearAxle+\mucarDreiLengthRefToFrontAxle}
\def\mucarDreiLengthComplete{\mucarDreiLengthRefToFrontEnd-\mucarDreiLengthRefToRearEnd}
\def\mucarDreiLengthRefToCenter{(\mucarDreiLengthRefToFrontEnd+\mucarDreiLengthRefToRearEnd)/2}
\def\mucarDreiLengthRearAxleToRearEnd{\mucarDreiLengthRefToRearEnd-\mucarDreiLengthRefToRearAxle}
\def\mucarDreiLengthRearAxleToFrontEnd{\mucarDreiLengthRefToFrontEnd-\mucarDreiLengthRefToRearAxle}
	
\def\scaling{3.0}
\def\RefLineSize{.125} 
	
\def\camera#1#2#3{
	\begin{scope}[shift={#1}, rotate=#2]
		\filldraw [fill=white,draw=#3](0,0) -- (\scaling * .15,\scaling * .15) -- (\scaling * -.15,\scaling * .15) -- cycle;
		\filldraw [fill=white,draw=#3](0,0) circle (\scaling * .075);
	\end{scope}
}

\def\cameraAngle#1#2#3#4{
	\begin{scope}[shift={#1}, rotate=#2]
		\filldraw [fill=white,draw=#3, fill=#3, fill opacity=0.3](0,0) -- ({tan(#4/2)*0.15*\scaling},.15*\scaling) -- ({-tan(#4/2)*0.15*\scaling},.15*\scaling) -- cycle;
		\filldraw [fill=white,draw=#3](0,0) circle (.075*\scaling);
		\filldraw [fill=#3, fill opacity=0.3,draw=#3](0,0) circle (.075 * \scaling);
	\end{scope}
}
\resizebox{.95\columnwidth}{!}{%
\begin{tikzpicture}
\node[inner sep=0pt, opacity=.1] (russell) at (0,0)
	{\includegraphics[width=42em]{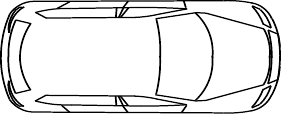}};

\coordinate (front_left_wheel)  at (\scaling * \mucarDreiLengthRefToFrontAxle,\scaling * \mucarDreiWidthRefToLeftFrontWheel);
\coordinate (front_right_wheel) at (\scaling * \mucarDreiLengthRefToFrontAxle,\scaling * \mucarDreiWidthRefToRightFrontWheel);
\coordinate (front_axle_center) at (\scaling * \mucarDreiLengthRefToFrontAxle, 0);
\coordinate (rear_left_wheel)   at (\scaling * \mucarDreiLengthRefToRearAxle, \scaling * \mucarDreiWidthRefToLeftRearWheel);
\coordinate (rear_right_wheel)  at (\scaling * \mucarDreiLengthRefToRearAxle, \scaling * \mucarDreiWidthRefToRightRearWheel);
\coordinate (rear_axle_center)  at (\mucarDreiLengthRefToRearAxle, 0);
\begin{scope}[inner sep=0, 
minimum width=\scaling * \mucarDreiReifendurchmesser cm,
minimum height=\scaling * \mucarDreiReifenbreite cm,
rounded corners=2pt]
	\node[draw,rectangle,at=(front_left_wheel)]{};
	\node[draw,rectangle,at=(front_right_wheel)]{};
	\node[draw,rectangle,at=(rear_left_wheel)]{};
	\node[draw,rectangle,at=(rear_right_wheel)]{};
\end{scope}

\draw[|-|, draw=black, thick] ([yshift=-2em]rear_right_wheel) -- ([yshift=-2em]front_right_wheel) node[midway,anchor=north] { \SI{2.855}{\meter}};

\draw[|-|, draw=black, thick] ([xshift=-.7*\mucarDreiReifendurchmesser*\scaling cm]rear_right_wheel) -- ([xshift=-.7*\mucarDreiReifendurchmesser*\scaling cm]rear_left_wheel) node[midway,anchor=east] {\SI{1.66}{\meter}};

\coordinate (front_bumper) at (\mucarDreiLengthRefToFrontEnd * \scaling + .25,-11em);
\coordinate (back_bumper) at (\mucarDreiLengthRefToRearEnd * \scaling - 0.2,-11em);
\draw[|-|, draw=black, thick] ([yshift=-.1em]front_bumper) -- ([yshift=-.1em]back_bumper) node[midway,anchor=north] { \SI{4.754}{\meter} };

\coordinate (surround_front) at (-0.243*\scaling + 0.2*\scaling,0 - 0.2*\scaling);
\coordinate (surround_front_beside) at (1.2*\scaling, 0 - 0.2*\scaling);
\coordinate (surround_front_center) at (1.2*\scaling, 0);
\cameraAngle{(surround_front)}{270}{blue}{94.55}
\draw[stealth-stealth, densely dashed, blue](surround_front_beside) -- (surround_front_center) node[midway,anchor=west]{\color{black}\SI{0.20}{\meter}};
\draw[-,thick,black] (surround_front_beside) ++(-\RefLineSize,0) -- ++(2*\RefLineSize,0);
\draw[-,thick,black] (surround_front_center) ++(-\RefLineSize,0) -- ++(2*\RefLineSize,0);

\coordinate (surround_right) at (-0.243*\scaling - 0.17*\scaling, 0 - 0.19*\scaling);
\cameraAngle{(surround_right)}{180}{blue}{94.55}
\coordinate (surround_right_beside) at (-0.243*\scaling - 0.17*\scaling, -.9*\scaling);
\coordinate (surround_right_center) at (0, -.9*\scaling);
\draw[stealth-stealth, densely dashed, blue](surround_right_beside) -- (surround_right_center) node[midway,anchor=north]{\color{black}\SI{0.42}{\meter}};
\draw[-,thick,black] (surround_right_beside) ++(0,-\RefLineSize) -- ++(0,2*\RefLineSize);
\draw[-,thick,black] (surround_right_center) ++(0,-\RefLineSize) -- ++(0,2*\RefLineSize); 

\coordinate (surround_left) at (-0.243*\scaling + 0.22*\scaling, 0 + 0.18*\scaling);
\cameraAngle{(surround_left)}{0}{blue}{94.55}
\coordinate (surround_left_beside) at (-0.243*\scaling + 0.22*\scaling, .48*\scaling);
\coordinate (surround_left_center) at (0, .48*\scaling);
\draw[-, densely dashed, blue](surround_left_beside) -- (surround_left_center) node[midway,anchor=west]{\color{black}\SI{0.02}{\meter}};
\draw[-,thick,black] (surround_left_beside) ++(0,-\RefLineSize) -- ++(0,2*\RefLineSize); 
\draw[-,thick,black] (surround_left_center) ++(0,-\RefLineSize) -- ++(0,2*\RefLineSize); 

\coordinate (surround_back) at (-0.243*\scaling - 0.2*\scaling, 0 + 0.15*\scaling);
\coordinate (surround_back_beside) at (-0.243*\scaling - 0.4*\scaling, 0 + 0.15*\scaling);
\coordinate (surround_back_center) at (-0.243*\scaling - 0.4*\scaling, 0);
\cameraAngle{(surround_back)}{90}{blue}{94.55}
\draw[stealth-stealth, densely dashed, blue](surround_back_beside) -- (surround_back_center) node[midway,anchor=east]{\color{black}\SI{0.15}{\meter}};
\draw[-,thick,black] (surround_back_beside) ++(-\RefLineSize,0) -- ++(2*\RefLineSize,0);
\draw[-,thick,black] (surround_back_center) ++(-\RefLineSize,0) -- ++(2*\RefLineSize,0);

\coordinate (windshield) at (-0.243*\scaling + 0.6*\scaling,0 - 0.15*\scaling);
\coordinate (windshield_beside) at (0.7*\scaling,-0.15*\scaling);
\coordinate (windshield_center) at (0.7*\scaling,0);
\cameraAngle{(windshield)}{270}{black!30!green}{60}
\draw[stealth-stealth, densely dashed, black!30!green](windshield_center) -- (windshield_beside) node[midway,anchor=west]{\color{black}\SI{0.15}{\meter}};
\draw[-,thick,black] (windshield_beside) ++(-\RefLineSize,0) -- ++(2*\RefLineSize,0);
\draw[-,thick,black] (windshield_center) ++(-\RefLineSize,0) -- ++(2*\RefLineSize,0);
\node[anchor=north west] at (windshield){\small RGB+NIR}; 

\coordinate (marveye_left) at (-0.243*\scaling + 0.67*\scaling + 0.02*\scaling, 0 + 0.03*\scaling + 0.06*\scaling);
\coordinate (marveye_left_beside) at (0.7*\scaling, 0 + 0.03*\scaling + 0.06*\scaling);
\coordinate (marveye_left_center) at (0.7*\scaling,0);
\cameraAngle{(marveye_left)}{270}{black!10!orange}{60}
\draw[stealth-stealth, densely dashed, {black!10!orange}](marveye_left_beside) -- (marveye_left_center) node[midway,anchor=west]{\color{black}\SI{0.10}{\meter}};
\draw[-,thick,black] (marveye_left_beside) ++(-\RefLineSize,0) -- ++(2*\RefLineSize,0);
\node[anchor=south] at (marveye_left){\small RGB}; 

\coordinate (ir_roof) at (-0.2*\scaling, 0);
\cameraAngle{(ir_roof)}{270}{black!10!pink}{60}
\node[anchor=south west] at (ir_roof){\small IR}; 

\coordinate (vls128_roof) at (-0.243* \scaling,0);
\coordinate (vls128_roof_beside) at (-0.243* \scaling,1.2* \scaling);
\coordinate (vls128_roof_center) at (0,1.2* \scaling);
\node[draw, circle, minimum size=\scaling*.2cm, red, at=(vls128_roof), fill=red, fill opacity=0.3]{};
\draw[stealth-stealth, densely dashed, red](vls128_roof_beside) -- (vls128_roof_center) node[midway,anchor=south]{\color{black}\SI{0.24}{\meter}};
\draw[-,thick,black] (vls128_roof_beside) ++(0,-\RefLineSize) -- ++(0,2*\RefLineSize);
\draw[-,thick,black] (vls128_roof_center) ++(0,-\RefLineSize) -- ++(0,2*\RefLineSize); 
\node[anchor=north] at (vls128_roof){\small LiDAR (128)}; 
\draw[-stealth, red](vls128_roof) -- ++(\scaling*0.2, 0) node[below,anchor=south] {}; 
\draw[-stealth, black!30!green](vls128_roof) -- ++(0, \scaling*0.2) node[below,anchor=south] {}; 

\coordinate (os32_left) at (\scaling*-0.243 + \scaling*0.03, 0 + \scaling*0.55);
\coordinate (os32_left_beside) at (\scaling*-0.243 -.85*\scaling, 0 + \scaling*0.55);
\coordinate (os32_left_center) at (\scaling*-0.243 -.85*\scaling, 0);
\draw[stealth-stealth, densely dashed, brown](os32_left_beside) -- (os32_left_center) node[midway,anchor=east]{\color{black}\SI{0.56}{\meter}};
\draw[-,thick,black] (os32_left_beside) ++(-\RefLineSize,0) -- ++(2*\RefLineSize,0);
\draw[-,thick,black] (os32_left_center) ++(-\RefLineSize,0) -- ++(2*\RefLineSize,0);
\node[draw, circle,minimum size = \scaling*0.2cm, inner sep=0pt, brown, at=(os32_left), fill=brown, fill opacity=0.3]{};
\coordinate (os32_left_x_beside) at (\scaling*-0.243 + \scaling*0.03,1.1*\scaling);
\coordinate (os32_left_x_center) at (0, 1.1*\scaling);
\draw[stealth-stealth, densely dashed, brown](os32_left_x_beside) -- (os32_left_x_center) node[midway,anchor=north]{\color{black}\SI{0.21}{\meter}};
\draw[-,thick,black] (os32_left_x_beside) ++(0,-\RefLineSize) -- ++(0,2*\RefLineSize);
\draw[-,thick,black] (os32_left_x_center) ++(0,-\RefLineSize) -- ++(0,2*\RefLineSize); 
\draw[-stealth, red](os32_left) -- ++(\scaling*0.2, 0) node[below,anchor=south] {x}; 
\draw[-stealth, blue, densely dotted](os32_left) -- ++(0, \scaling*0.2*0.707) node[below,anchor=south] {z};
\draw[-stealth, black!30!green, densely dotted]([xshift=0.05em]os32_left) -- ++(0., \scaling*0.2*0.707) node[below,anchor=east] {y}; 
\node[anchor=east] at (os32_left){\small $\phi = -45^{\circ}$};
\node[anchor=north east] at (os32_left){\small LiDAR (32)}; 

\coordinate (os32_right) at (\scaling*-0.243 + \scaling*0.07, 0 - \scaling*0.55);
\coordinate (os32_right_beside) at (\scaling*-0.243 -.85*\scaling, 0 - \scaling*0.55);
\coordinate (os32_right_center) at (\scaling*-0.243 -.85*\scaling, 0);
\node[draw, circle,minimum size = \scaling*0.2cm, inner sep=0pt, brown, at=(os32_right), fill=brown, fill opacity=0.3]{};
\draw[stealth-stealth, densely dashed, brown](os32_right_beside) -- (os32_right_center) node[midway,anchor=east]{\color{black}\SI{0.56}{\meter}};
\draw[-,thick,black] (os32_right_center) ++(-\RefLineSize,0) -- ++(2*\RefLineSize,0);
\draw[-,thick,black] (os32_right_beside) ++(-\RefLineSize,0) -- ++(2*\RefLineSize,0);
\draw[-stealth, red](os32_right) -- ++(\scaling*0.2, 0) node[below,anchor=south] {x}; 
\draw[-stealth, blue, densely dotted](os32_right) -- ++(0, -\scaling*0.2*0.707) node[below,anchor=north] {z};
\draw[-stealth, black!30!green, densely dotted](os32_right) -- ++(0., +\scaling*0.2*0.707) node[below,anchor=east] {y}; 
\node[anchor=east] at (os32_right){\small $\phi = 45^{\circ}$};
\node[anchor=north east] at (os32_right){\small LiDAR (32)}; 

\coordinate (umrr_front_left) at ( \scaling*2.155 -  \scaling*0.1,  \scaling*0.8+ \scaling*0.1 );
\coordinate (umrr_back_front_beside) at (\scaling*2.155 -  \scaling*0.1, \scaling*1.4);
\coordinate (umrr_back_front_center) at (0, \scaling*1.4);
\draw[stealth-stealth, densely dashed, pink](umrr_back_front_beside) -- (umrr_back_front_center) node[midway,anchor=south]{\color{black}\SI{2.06}{\meter}};
\draw[-,thick,black] (umrr_back_front_beside) ++(0,-\RefLineSize) -- ++(0,2*\RefLineSize); 
\draw[-,thick,black] (umrr_back_front_center) ++(0,-\RefLineSize) -- ++(0,2*\RefLineSize); 
\node[draw, rectangle,minimum size =  \scaling*0.15cm, inner sep=0pt, pink, at=(umrr_front_left), fill=pink, fill opacity=0.3]{};
\draw[-stealth, red](umrr_front_left) -- ++({0.2*\scaling*sin(51)}, {0.2*\scaling*cos(51)})
node[below,anchor=south] {x};
\draw[-stealth, black!30!green](umrr_front_left) -- ++({-0.2*\scaling*sin(141)}, {-0.2*\scaling*cos(141)}) node[below,anchor=east] {y};
\node[anchor=north] at (umrr_front_left){\small SRR};

\coordinate (umrr_front_right) at (\scaling*2.155 - \scaling*0.1, \scaling*-0.8 + \scaling*-0.1 );
\node[draw, rectangle,minimum size = \scaling*0.15cm, inner sep=0pt, pink, at=(umrr_front_right), fill=pink, fill opacity=0.3]{};
\draw[-stealth, red](umrr_front_right) -- ++(-{0.2*\scaling*sin(-46)}, {-0.2*\scaling*cos(-46)})
node[below,anchor=south] {x};
\draw[-stealth, black!30!green](umrr_front_right) -- ++({-0.2*\scaling*sin(44)}, {-0.2*\scaling*cos(44)}) node[below,anchor=east] {y};
\node[anchor=south] at (umrr_front_right){\small SRR};

\coordinate (umrr_front_center_left) at (\scaling*2.155 + \scaling*0.15, \scaling*0.195 );
\coordinate (umrr_front_center_left_beside) at (\scaling*2.155 + \scaling*0.15, \scaling*1.4);
\coordinate (umrr_front_center_left_center) at (\scaling*2.155 -  \scaling*0.1, \scaling*1.4);
\draw[stealth-stealth, densely dashed, pink](umrr_front_center_left_beside) -- (umrr_front_center_left_center) node[midway,anchor=south]{\color{black}\SI{0.09}{\meter}};
\draw[-,thick,black] (umrr_front_center_left_beside) ++(0,-\RefLineSize) -- ++(0,2*\RefLineSize); 
\node[draw, rectangle,minimum size = \scaling*0.15cm, inner sep=0pt, pink, at=(umrr_front_center_left), fill=pink, fill opacity=0.3]{};
\draw[-stealth, red](umrr_front_center_left) -- ++({0.2*\scaling*sin(91)}, {0.2*\scaling*cos(91)})
node[below,anchor=south] {x};
\draw[-stealth, black!30!green](umrr_front_center_left) -- ++({0.2*\scaling*sin(1)}, {0.2*\scaling*cos(1)}) node[below,anchor=east] {y};
\node[anchor=east] at (umrr_front_center_left){\small FRR};

\coordinate (umrr_front_center_right) at (\scaling*2.155 + \scaling*0.15, \scaling*-0.121 );
\coordinate (umrr_front_center_right_beside) at (\scaling*2.6, \scaling*-0.121 );
\coordinate (umrr_front_center_right_center) at (\scaling*2.6, 0 );
\node[draw, rectangle,minimum size = \scaling*0.15cm, inner sep=0pt, pink, at=(umrr_front_center_right), fill=pink, fill opacity=0.3]{};
\draw[stealth-stealth, densely dashed, pink](umrr_front_center_right_beside) -- (umrr_front_center_right_center) node[midway,anchor=west]{\color{black}\SI{0.19}{\meter}};
\draw[-,thick,black] (umrr_front_center_right_center) ++(-\RefLineSize,0) -- ++(2*\RefLineSize,0);
\draw[-,thick,black] (umrr_front_center_right_beside) ++(-\RefLineSize,0) -- ++(2*\RefLineSize,0);  
\draw[-stealth, red](umrr_front_center_right) -- ++({0.2*\scaling*sin(89)}, {0.2*\scaling*cos(89)})
node[below,anchor=south] {x};
\draw[-stealth, black!30!green](umrr_front_center_right) -- ++({0.2*\scaling*sin(-0.49)}, {0.2*\scaling*cos(-0.49)}) node[below,anchor=east] {y};
\node[anchor=east] at (umrr_front_center_right){\small SRR};

\coordinate (umrr_back_left) at (\scaling*-0.243 - \scaling*1.9 + \scaling*0.1, \scaling*0.8+\scaling*0.1);
\coordinate (umrr_back_left_beside) at (\scaling*-0.243 - \scaling*1.9 + \scaling*0.1, \scaling*1.4);
\coordinate (umrr_back_left_center) at (0, \scaling*1.4);
\coordinate (umrr_back_left_y_beside) at (\scaling*-2.5, \scaling*0.8+\scaling*0.1);
\coordinate (umrr_back_left_y_center) at (\scaling*-2.5, 0);
\node[draw, rectangle,minimum size = \scaling*0.15cm, inner sep=0pt, pink, at=(umrr_back_left), fill=pink, fill opacity=0.3]{};
\draw[stealth-stealth, densely dashed, pink](umrr_back_left_beside) -- (umrr_back_left_center) node[midway,anchor=south]{\color{black}\SI{2.14}{\meter}};
\draw[-,thick,black] (umrr_back_left_beside) ++(0,-\RefLineSize) -- ++(0,2*\RefLineSize); 
\draw[-,thick,black] (umrr_back_left_center) ++(0,-\RefLineSize) -- ++(0,2*\RefLineSize); 
\draw[stealth-stealth, densely dashed, pink](umrr_back_left_y_beside) -- (umrr_back_left_y_center) node[midway,anchor=east]{\color{black}\SI{0.8}{\meter}};
\draw[-,thick,black] (umrr_back_left_y_center) ++(-\RefLineSize,0) -- ++(2*\RefLineSize,0);
\draw[-,thick,black] (umrr_back_left_y_beside) ++(-\RefLineSize,0) -- ++(2*\RefLineSize,0);
\draw[-stealth, red](umrr_back_left) -- ++({-0.2*\scaling*sin(142)}, {-0.2*\scaling*cos(142)})
node[below,anchor=south] {x};
\draw[-stealth, black!30!green](umrr_back_left) -- ++({0.2*\scaling*sin(232)}, {0.2*\scaling*cos(232)}) node[below,anchor=east] {y};
\node[anchor=west] at (umrr_back_left){\small SRR};

\coordinate (umrr_back_right) at (\scaling*-0.243 - \scaling*1.9 + \scaling*0.1, \scaling*-0.8 +\scaling*-0.1);
\coordinate (umrr_back_right_beside) at (\scaling*-2.5, \scaling*-0.8 +\scaling*-0.1);
\coordinate (umrr_back_right_center) at (\scaling*-2.5, 0);
\node[draw, rectangle,minimum size = \scaling*0.15cm, inner sep=0pt, pink, at=(umrr_back_right), fill=pink, fill opacity=0.3]{};
\draw[stealth-stealth, densely dashed, pink](umrr_back_right_beside) -- (umrr_back_right_center) node[midway,anchor=east]{\color{black}\SI{0.8}{\meter}};
\draw[-,thick,black] (umrr_back_right_beside) ++(-\RefLineSize,0) -- ++(2*\RefLineSize,0);
\draw[-,thick,black] (umrr_back_right_center) ++(-\RefLineSize,0) -- ++(2*\RefLineSize,0);
\draw[-stealth, red](umrr_back_right) -- ++({0.2*\scaling*sin(-137)}, {0.2*\scaling*cos(-137)})
node[below,anchor=south] {x};
\draw[-stealth, black!30!green](umrr_back_right) -- ++({-0.2*\scaling*sin(-227)}, {-0.2*\scaling*cos(-227)}) node[below,anchor=east] {y};
\node[anchor=west] at (umrr_back_right){\small SRR};

\coordinate (base_link) at (0, 0);
\draw[-stealth, red](base_link) -- (\scaling*0.3, 0) node[below,anchor=south] {x}; 
\draw[-stealth, black!30!green](base_link) -- (0, \scaling*0.3) node[below,anchor=south] {y}; 
\node[draw, circle, fill,minimum size = 0.1cm, inner sep=.5pt, black, at=(base_link)]{};
	
\end{tikzpicture}
}

%% file: images/sensor_setup/side_view.tex
\def\mucarDreiReifendurchmesser{.737}
\def\mucarDreiReifenbreite{.255}
\def\mucarDreiWidth{1.928}
\def\mucarDreiLengthRefToFrontEnd{2.3535}
\def\mucarDreiLengthRefToRearEnd{-2.4005}
\def\mucarDreiLengthRefToFrontAxle{1.4275}
\def\mucarDreiLengthRefToRearAxle{-1.4275}
\def\mucarDreiWidthRefToLeftFrontWheel{.8265}
\def\mucarDreiWidthRefToRightFrontWheel{-.8265}
\def\mucarDreiWidthRefToLeftRearWheel{.83225}
\def\mucarDreiWidthRefToRightRearWheel{-.83225}
\def\mucarThreeWheelbase{-\mucarDreiLengthRefToRearAxle+\mucarDreiLengthRefToFrontAxle}
\def\mucarDreiLengthComplete{\mucarDreiLengthRefToFrontEnd-\mucarDreiLengthRefToRearEnd}
\def\mucarDreiLengthRefToCenter{(\mucarDreiLengthRefToFrontEnd+\mucarDreiLengthRefToRearEnd)/2}
\def\mucarDreiLengthRearAxleToRearEnd{\mucarDreiLengthRefToRearEnd-\mucarDreiLengthRefToRearAxle}
\def\mucarDreiLengthRearAxleToFrontEnd{\mucarDreiLengthRefToFrontEnd-\mucarDreiLengthRefToRearAxle}
	
\def\scaling{3.0}

\def\ground{-1.1} 
\def\GroundToRef{0.993}
\def\RefToVls{1.5} 
\def\VlsPos{\ground + \GroundToRef + \RefToVls}
\def\RefLineSize{.125}

\def\camera#1#2#3{
	\begin{scope}[shift={#1}, rotate=#2]
		\filldraw [fill=white,draw=#3](0,0) -- (.15 * \scaling,.15 * \scaling) -- (-.15 * \scaling,.15 * \scaling) -- cycle;
		\filldraw [fill=white,draw=#3](0,0) circle (.075 * \scaling);
	\end{scope}
}

\def\cameraAngle#1#2#3#4{
	\begin{scope}[shift={#1}, rotate=#2]
		\filldraw [fill=white,draw=#3, fill=#3, fill opacity=0.3](0,0) -- ({tan(#4/2)*0.15 * \scaling},.15 * \scaling) -- ({-tan(#4/2)*0.15 * \scaling},.15 * \scaling) -- cycle;
		\filldraw [fill=white,draw=#3](0,0) circle (.075 * \scaling);
		\filldraw [fill=#3, fill opacity=0.3,draw=#3](0,0) circle (.075 * \scaling);
	\end{scope}
}

\resizebox{.95\columnwidth}{!}{%
\begin{tikzpicture}
\node[inner sep=0pt] (russell) at (0,0)
	{\includegraphics[width=42em, 
		trim=10 10 10 10, clip]{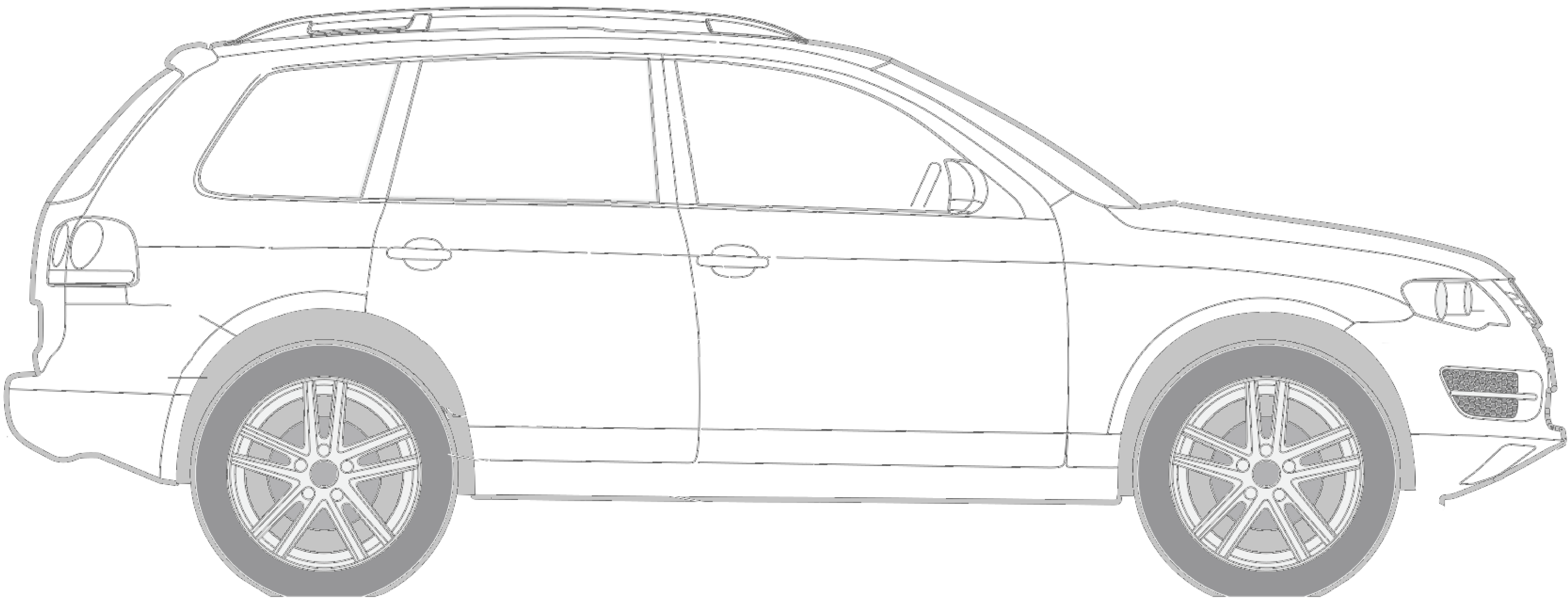}};

\coordinate (front_right_wheel) at (\mucarDreiLengthRefToFrontAxle * \scaling,\mucarDreiWidthRefToRightFrontWheel * \scaling);
\coordinate (rear_right_wheel)  at (\mucarDreiLengthRefToRearAxle * \scaling, \mucarDreiWidthRefToRightFrontWheel * \scaling);

\coordinate (front_bumper) at (\mucarDreiLengthRefToFrontEnd * \scaling + .25,\ground * \scaling);
\coordinate (back_bumper) at (\mucarDreiLengthRefToRearEnd * \scaling,\ground * \scaling);

\draw[|-|, draw=black, thick] ([yshift=-1.1em]rear_right_wheel) -- ([yshift=-1.1em]front_right_wheel) node[midway,anchor=north] { \SI{2.855}{\meter} };
\draw[|-|, draw=black, thick] ([xshift=-1em]\mucarDreiLengthRefToRearEnd * \scaling,-\mucarDreiWidthRefToRightRearWheel * \scaling) -- ([xshift=-1em]\mucarDreiLengthRefToRearEnd * \scaling,\mucarDreiWidthRefToRightRearWheel * \scaling) node[midway,anchor=east] {  \SI{1.72}{\meter} };

\draw[|-|, draw=black, thick] ([yshift=-.1em]front_bumper) -- ([yshift=-.1em]back_bumper) node[midway,anchor=north] { \SI{4.754}{\meter} };

\coordinate (surround_front) at (-0.243 * \scaling + 0.2 * \scaling,\VlsPos * \scaling - 0.13 * \scaling);
\cameraAngle{(surround_front)}{270}{blue}{46.5}

\coordinate (surround_back) at (-0.243 * \scaling - 0.2 * \scaling,\VlsPos * \scaling - 0.13 * \scaling);
\coordinate (beside_surround_back) at (-0.243 * \scaling - 0.6 * \scaling,\VlsPos * \scaling - 0.13 * \scaling);
\coordinate (bottom_surround_back) at (-0.243 * \scaling - 0.6 * \scaling,\ground * \scaling + \GroundToRef * \scaling);
\draw[-,thick,black] (beside_surround_back) ++(-\RefLineSize,0) -- ++(2*\RefLineSize,0); 
\draw[-,thick,black] (bottom_surround_back) ++(-\RefLineSize,0) -- ++(2*\RefLineSize,0);
\draw[stealth-stealth,densely dashed, thick,blue](beside_surround_back) -- (bottom_surround_back) node[midway, anchor=east]{\color{black} \SI{1.46}{\meter}};
\node[anchor=east] at (beside_surround_back){\small RGB};

\cameraAngle{(surround_back)}{90}{blue}{46.5}

\coordinate (windshield) at (-0.243 * \scaling + 0.6 * \scaling ,\VlsPos * \scaling - 0.64 * \scaling - 0.2 * \scaling);
\coordinate (beside_windshield) at (-0.243 * \scaling + 0.6 * \scaling + 0.2 * \scaling,\VlsPos * \scaling - 0.64 * \scaling - 0.2 * \scaling);
\coordinate (bottom_windshield) at (-0.243 * \scaling + 0.6 * \scaling + 0.2 * \scaling,\ground * \scaling + \GroundToRef * \scaling);
\draw[-,thick,black] (beside_windshield) ++(-\RefLineSize,0) -- ++(2*\RefLineSize,0); 
\draw[-,thick,black] (bottom_windshield) ++(-\RefLineSize,0) -- ++(2*\RefLineSize,0); 
\node[anchor=west] at (beside_windshield){\small RGB+NIR};
\draw[stealth-stealth,densely dashed, thick,black!30!green](beside_windshield) -- (bottom_windshield) node[midway, anchor=west]{\color{black} \SI{0.95}{\meter}};

\cameraAngle{(windshield)}{270}{black!30!green}{46}


\coordinate (ir_roof) at (-0.2 * \scaling, \ground * \scaling + \GroundToRef * \scaling + 1.245 * \scaling);
\coordinate (ir_roof_beside) at (-0.2 * \scaling + 0.3 * \scaling, \ground * \scaling + \GroundToRef * \scaling + 1.245 * \scaling);
\coordinate (ir_roof_bottom) at (-0.2 * \scaling + 0.3 * \scaling, \ground * \scaling + \GroundToRef * \scaling);

\draw[-,thick,black] (ir_roof_beside) ++(-\RefLineSize,0) -- ++(2*\RefLineSize,0); 
\draw[-,thick,black] (ir_roof_bottom) ++(-\RefLineSize,0) -- ++(2*\RefLineSize,0);
\node[anchor=west] at (ir_roof_beside){\small IR};
\draw[stealth-stealth,densely dashed, thick,black!10!pink](ir_roof_beside) -- (ir_roof_bottom) node[midway, anchor=west]{\color{black} \SI{1.245}{\meter}};
\cameraAngle{(ir_roof)}{270}{black!10!pink}{60}

\coordinate (vls128_roof) at (-0.243 * \scaling,\VlsPos * \scaling);
\coordinate (vls128_roof_beside) at (-0.33 * \scaling,\VlsPos * \scaling);
\coordinate (vls128_roof_bottom) at (-0.33 * \scaling,\ground * \scaling + \GroundToRef * \scaling);
\node[draw, rectangle, fill=red, fill opacity=0.3, minimum size=\scaling * .2cm, red, at=(vls128_roof)]{};
\draw[-,thick,black] (vls128_roof_beside) ++(-\RefLineSize,0) -- ++(2*\RefLineSize,0); 
\draw[-,thick,black] (vls128_roof_bottom) ++(-\RefLineSize,0) -- ++(2*\RefLineSize,0);
\node[anchor=east] at (vls128_roof_beside){\small LiDAR (128)};
\draw[stealth-stealth,densely dashed,red](vls128_roof_beside) -- (vls128_roof_bottom) node[midway, anchor=west]{\color{black} \SI{1.6}{\meter}};
\draw[-stealth, red](vls128_roof) -- ++(0.3,0) node[below,anchor=north] {}; 
\draw[-stealth, blue](vls128_roof) -- ++(0,0.3) node[anchor=west] {}; 

\coordinate (os32_left) at (-0.243 * \scaling - 0.03 * \scaling, \VlsPos * \scaling - 0.37 * \scaling -0.15 * \scaling);
\coordinate (os32_left_beside) at (-0.243 * \scaling -.2 * \scaling, \VlsPos * \scaling - 0.37 * \scaling -0.15 * \scaling);
\coordinate (os32_left_bottom) at (-0.243 * \scaling -.2 * \scaling,\ground * \scaling + \GroundToRef * \scaling);
\node[draw, rectangle, fill=brown, fill opacity=0.3,minimum size = \scaling * 0.2cm, inner sep=0pt, brown, at=(os32_left)]{};
\draw[-,thick,black] (os32_left_beside) ++(-\RefLineSize,0) -- ++(2*\RefLineSize,0); 
\draw[-,thick,black] (os32_left_bottom) ++(-\RefLineSize,0) -- ++(2*\RefLineSize,0);
\node[anchor=south east] at (os32_left){\small LiDAR (32)};
\draw[stealth-stealth,densely dashed,brown](os32_left_beside) -- (os32_left_bottom) node[midway, anchor=east]{\color{black} \SI{1.23}{\meter}};
\draw[-stealth, red](os32_left) -- ++(0.3,0) node[below,anchor=north] {}; 
\draw[-stealth, blue](os32_left) -- ++(0,0.3) node[anchor=west] {}; 

\coordinate (umrr_front) at (2.155 * \scaling, \ground * \scaling + \GroundToRef * \scaling);
\coordinate (beside_umrr_front) at  (2.155 * \scaling + 0.15 * \scaling, \ground * \scaling + \GroundToRef * \scaling);
\coordinate (bottom_umrr_front) at  (2.155 * \scaling + 0.15 * \scaling, \ground * \scaling + 0.25 * \scaling);
\draw[-,thick,black] (beside_umrr_front) ++(-\RefLineSize,0) -- ++(2*\RefLineSize,0); 
\draw[-,thick,black] (bottom_umrr_front) ++(-\RefLineSize,0) -- ++(2*\RefLineSize,0);
\node[draw, rectangle,minimum size = \scaling * 0.15cm, inner sep=0pt, pink, at=(umrr_front), fill=pink, fill opacity=0.3]{};
\draw[stealth-stealth,densely dashed,pink](beside_umrr_front) -- (bottom_umrr_front) node[midway, anchor=east]{\color{black} \SI{0.96}{\meter}};
\draw[-stealth, red](umrr_front) -- ++(0.3,0) node[below,anchor=north] {}; 
\draw[-stealth, blue](umrr_front) -- ++(0,-0.3) node[anchor=west] {}; 
\node[anchor=north east] at (umrr_front){\small SRR};

\coordinate (umrr_back) at (-0.243 * \scaling - 1.9 * \scaling - 0.1 * \scaling, \ground * \scaling + \GroundToRef * \scaling);
\coordinate (beside_umrr_back) at  (-0.243 * \scaling - 1.9 * \scaling, \ground * \scaling + \GroundToRef * \scaling);
\coordinate (bottom_umrr_back) at  (-0.243 * \scaling - 1.9 * \scaling, \ground * \scaling + 0.25 * \scaling);
\draw[-,thick,black] (beside_umrr_back) ++(-\RefLineSize,0) -- ++(2*\RefLineSize,0); 
\draw[-,thick,black] (bottom_umrr_back) ++(-\RefLineSize,0) -- ++(2*\RefLineSize,0);
\node[draw, rectangle,minimum size = \scaling * 0.15cm, inner sep=0pt, pink, at=(umrr_back), fill=pink, fill opacity=0.3]{};
\draw[stealth-stealth,densely dashed,pink](beside_umrr_back) -- (bottom_umrr_back) node[midway, anchor=east]{\color{black} \SI{0.96}{\meter}};
\draw[-stealth, red](umrr_back) -- ++(-0.3,0) node[below,anchor=north] {}; 
\draw[-stealth, blue](umrr_back) -- ++(0,0.3) node[anchor=west] {}; 
\node[anchor=north west] at (umrr_back){\small SRR};

\coordinate (umrr_front_hood) at (2.155 * \scaling + 0.25 * \scaling, \ground * \scaling + \GroundToRef * \scaling - 0.1 * \scaling) ;
\coordinate (beside_umrr_front_hood) at  (2.155 * \scaling + 0.35 * \scaling, \ground * \scaling + \GroundToRef * \scaling - 0.1 * \scaling);
\coordinate (bottom_umrr_front_hood) at  (2.155 * \scaling + 0.35 * \scaling, \ground * \scaling + 0.25 * \scaling);
\draw[stealth-stealth,densely dashed,pink](beside_umrr_front_hood) -- (bottom_umrr_front_hood) node[midway, anchor=west]{\color{black} \SI{0.84}{\meter}};
\draw[-,thick,black] (beside_umrr_front_hood) ++(-\RefLineSize,0) -- ++(2*\RefLineSize,0); 
\draw[-,thick,black] (bottom_umrr_front_hood) ++(-\RefLineSize,0) -- ++(2*\RefLineSize,0);
\node[draw, rectangle,minimum size = \scaling * 0.15cm, inner sep=0pt, pink, at=(umrr_front_hood), fill=pink, fill opacity=0.3]{};
\draw[-stealth, red](umrr_front_hood) -- ++(0.3,0) node[below,anchor=north] {}; 
\draw[-stealth, blue](umrr_front_hood) -- ++(0,0.3) node[anchor=west] {}; 
\node[anchor=north west] at (umrr_front_hood){\small FRR};

\coordinate (base_link) at (0, \ground * \scaling + \GroundToRef * \scaling);
\draw[-stealth, red](base_link) -- (0.3 * \scaling, \ground * \scaling + \GroundToRef * \scaling) node[below,anchor=north] {x}; 
\draw[-stealth, blue](base_link) -- (0.0, \ground * \scaling + \GroundToRef * \scaling + 0.3 * \scaling) node[anchor=west] { z }; 
\node[draw, circle, fill,minimum size = \scaling * 0.01cm, inner sep=.5pt, black, at=(base_link)]{};

\coordinate (base_link_height) at (-.7 * \scaling, \ground * \scaling + \GroundToRef * \scaling);
\coordinate (base_link_ground) at (-.7 * \scaling, \ground * \scaling + 0.25 * \scaling);
\draw[-,thick,black] (base_link_height) ++(-\RefLineSize,0) -- ++(2*\RefLineSize,0); 
\draw[-,thick,black] (base_link_ground) ++(-\RefLineSize,0) -- ++(2*\RefLineSize,0);
\draw[stealth-stealth,densely dashed, black] (base_link_height) -- (base_link_ground) node[midway, anchor=south west] { \SI{0.64}{\meter}};

\node[draw, rectangle, fill=black, fill opacity=0.1, minimum size = \scaling * 0.15cm, inner sep=0pt] at (base_link){};
\node[anchor=north west] at (base_link){\small INS};
	
\end{tikzpicture}
}

%% file: root.bbl
\begin{thebibliography}{10}
\providecommand{\url}[1]{#1}
\csname url@samestyle\endcsname
\providecommand{\newblock}{\relax}
\providecommand{\bibinfo}[2]{#2}
\providecommand{\BIBentrySTDinterwordspacing}{\spaceskip=0pt\relax}
\providecommand{\BIBentryALTinterwordstretchfactor}{4}
\providecommand{\BIBentryALTinterwordspacing}{\spaceskip=\fontdimen2\font plus
\BIBentryALTinterwordstretchfactor\fontdimen3\font minus
  \fontdimen4\font\relax}
\providecommand{\BIBforeignlanguage}[2]{{%
\expandafter\ifx\csname l@#1\endcsname\relax
\typeout{** WARNING: IEEEtran.bst: No hyphenation pattern has been}%
\typeout{** loaded for the language `#1'. Using the pattern for}%
\typeout{** the default language instead.}%
\else
\language=\csname l@#1\endcsname
\fi
#2}}
\providecommand{\BIBdecl}{\relax}
\BIBdecl

\bibitem{neigel_offsed_2021}
P.~Neigel, J.~Rambach, and D.~Stricker, ``{OFFSED: Off-Road Semantic
  Segmentation Dataset},'' in \emph{{Proceedings of International Conference on
  Computer Vision Theory and Applications (VISAPP)}}, 2021.

\bibitem{valada_deepscene_2016}
A.~Valada, G.~Oliveira, T.~Brox, and W.~Burgard, ``{Deep Multispectral Semantic
  Scene Understanding of Forested Environments using Multimodal Fusion},'' in
  \emph{{International Symposium on Experimental Robotics (ISER)}}, 2016.

\bibitem{metzger_tas500_2020}
K.~A. Metzger, P.~Mortimer, and H.-J. Wuensche, ``{A Fine-Grained Dataset and
  its Efficient Semantic Segmentation for Unstructured Driving Scenarios},'' in
  \emph{{International Conference on Pattern Recognition (ICPR)}}, 2021.

\bibitem{maturana_ycor_2018}
D.~Maturana, P.-W. Chou, M.~Uenoyama, and S.~Scherer, ``{Real-Time Semantic
  Mapping for Autonomous Off-Road Navigation},'' in \emph{Field and Service
  Robotics}.\hskip 1em plus 0.5em minus 0.4em\relax Springer, 2018.

\bibitem{wigness_rugd_2019}
M.~Wigness, S.~Eum, J.~G. Rogers, D.~Han, and H.~Kwon, ``A {RUGD} {Dataset} for
  {Autonomous} {Navigation} and {Visual} {Perception} in {Unstructured}
  {Outdoor} {Environments},'' in \emph{{Proceedings of IEEE/RSJ International
  Conference on Intelligent Robots and Systems (IROS)}}, Macau, China, 2019.

\bibitem{jiang_rellis-3d_2021}
P.~Jiang, P.~Osteen, M.~Wigness, and S.~Saripalli, ``{RELLIS}-{3D} {Dataset}:
  {Data}, {Benchmarks} and {Analysis},'' in \emph{{Proceedings of IEEE
  International Conference on Robotics and Automation (ICRA)}}, Xi'an, China,
  May 2021.

\bibitem{geiger_vision_2013}
A.~Geiger, P.~Lenz, C.~Stiller, and R.~Urtasun, ``{Vision meets Robotics: The
  KITTI Dataset},'' \emph{{The International Journal of Robotics Research}},
  vol.~32, no.~11, 2013.

\bibitem{cordts_cityscapes_2016}
M.~Cordts, M.~Omran, S.~Ramos, T.~Rehfeld, M.~Enzweiler, R.~Benenson,
  U.~Franke, S.~Roth, and B.~Schiele, ``The {Cityscapes} {Dataset} for
  {Semantic} {Urban} {Scene} {Understanding},'' in \emph{{Proceedings of IEEE
  Conference on Computer Vision and Pattern Recognition (CVPR)}}, Las Vegas,
  NV, USA, 2016.

\bibitem{caesar_nuscenes_2020}
H.~Caesar, V.~Bankiti, A.~H. Lang, S.~Vora, V.~E. Liong, Q.~Xu, A.~Krishnan,
  Y.~Pan, G.~Baldan, and O.~Beijbom, ``{nuScenes}: {A} {Multimodal} {Dataset}
  for {Autonomous} {Driving},'' in \emph{{Proceedings of IEEE Conference on
  Computer Vision and Pattern Recognition (CVPR)}}, 2020.

\bibitem{behley_semantickitti_2019}
J.~Behley, M.~Garbade, A.~Milioto, J.~Quenzel, S.~Behnke, C.~Stachniss, and
  J.~Gall, ``{SemanticKITTI}: {A} {Dataset} for {Semantic} {Scene}
  {Understanding} of {LiDAR} {Sequences},'' in \emph{{Proceedings of IEEE
  International Conference on Computer Vision (ICCV)}}, Seoul, Korea (South),
  2019.

\bibitem{zendel_wilddash2_2022}
O.~Zendel, M.~Sch\"orghuber, B.~Rainer, M.~Murschitz, and C.~Beleznai,
  ``{Unifying Panoptic Segmentation for Autonomous Driving},'' in
  \emph{{Proceedings of IEEE Conference on Computer Vision and Pattern
  Recognition (CVPR)}}, 2022.

\bibitem{min_orfd_2022}
C.~Min, W.~Jiang, D.~Zhao, J.~Xu, L.~Xiao, Y.~Nie, and B.~Dai, ``{ORFD: A
  Dataset and Benchmark for Off-Road Freespace Detection},'' in
  \emph{{Proceedings of IEEE International Conference on Robotics and
  Automation (ICRA)}}, 2022, pp. 2532--2538.

\bibitem{knights_wildplaces_2023}
J.~Knights, K.~Vidanapathirana, M.~Ramezani, S.~Sridharan, C.~Fookes, and
  P.~Moghadam, ``{Wild-Places: A Large-Scale Dataset for Lidar Place
  Recognition in Unstructured Natural Environments},'' in \emph{{Proceedings of
  IEEE International Conference on Robotics and Automation (ICRA)}}, 2023.

\bibitem{triest_tartandrive_2022}
S.~Triest, M.~Sivaprakasam, S.~J. Wang, W.~Wang, A.~M. Johnson, and S.~Scherer,
  ``{TartanDrive: A Large-Scale Dataset for Learning Off-Road Dynamics
  Models},'' in \emph{{Proceedings of IEEE International Conference on Robotics
  and Automation (ICRA)}}, 2022.

\bibitem{tampuu_estoniadriving_2023}
A.~Tampuu, R.~Aidla, J.~A. van Gent, and T.~Matiisen, ``{LiDAR-as-Camera for
  End-to-End Driving},'' \emph{Sensors}, vol.~23, no.~5, 2023.

\bibitem{smith_atlas_2022}
W.~Smith, D.~Grabowsky, and D.~Mikulski, ``{ATLAS, an All-Terrain Lalbelset for
  Autonomous Systems},'' in \emph{Ground Vehicle Systems Engineering and
  Technology Symposium}, 2022.

\bibitem{lambert_mseg_2020}
J.~Lambert, Z.~Liu, O.~Sener, J.~Hays, and V.~Koltun, ``{MSeg: A Composite
  Dataset for Multi-domain Semantic Segmentation},'' in \emph{{Proceedings of
  IEEE Conference on Computer Vision and Pattern Recognition (CVPR)}}, 2020.

\bibitem{caesar_cocostuff_2018}
H.~Caesar, J.~Uijlings, and V.~Ferrari, ``{COCO-Stuff: Thing and Stuff Classes
  in Context},'' in \emph{{Proceedings of IEEE Conference on Computer Vision
  and Pattern Recognition (CVPR)}}, 2018.

\bibitem{quigley_ros1_2009}
M.~Quigley, B.~Gerkey, K.~Conley, J.~Faust, T.~Foote, J.~Leibs, E.~Berger,
  R.~Wheeler, and A.~Ng, ``{ROS: an open-source Robot Operating System},'' in
  \emph{{Proceedings of IEEE International Conference on Robotics and
  Automation Workshops (ICRAW)}}, 2009.

\bibitem{himmelsbach_mucar3_2011}
M.~Himmelsbach, T.~Luettel, F.~Hecker, F.~{v. Hundelshausen}, and H.-J.
  Wuensche, ``{Autonomous Off-Road Navigation for MuCAR-3},''
  \emph{KI-K{\"u}nstliche Intelligenz}, vol.~25, no.~2, 2011.

\bibitem{unterholzner_marveye_2010}
A.~Unterholzner and H.-J. Wuensche, ``{Hybrid Adaptive Control of a Multi-Focal
  Vision System},'' in \emph{{IEEE} Transactions on Intelligent Vehicles},
  2010.

\bibitem{ieee_ptp}
``{IEEE Standard for a Precision Clock Synchronization Protocol for Networked
  Measurement and Control Systems},'' \emph{IEEE Std 1588-2019 (Revision of
  IEEE Std 1588-2008)}, pp. 1--499, 2020.

\bibitem{wang_apriltag2_2016}
J.~Wang and E.~Olson, ``{AprilTag 2: Efficient and robust fiducial
  detection},'' in \emph{{Proceedings of IEEE/RSJ International Conference on
  Intelligent Robots and Systems (IROS)}}, 2016.

\bibitem{kallwies_apriltagcorners_2020}
J.~Kallwies, B.~Forkel, and H.-J. Wuensche, ``{Determining and Improving the
  Localization Accuracy of AprilTag Detection},'' in \emph{{Proceedings of IEEE
  International Conference on Robotics and Automation (ICRA)}}, 2020.

\bibitem{guindel_lidarcalibrationboard_2017}
C.~Guindel, J.~Beltrán, D.~Martín, and F.~García, ``{Automatic Extrinsic
  Calibration for Lidar-Stereo Vehicle Sensor Setups},'' in \emph{{Proceedings
  of IEEE Intelligent Transportation Systems Conference (ITSC)}}, 2017.

\bibitem{beltran_lidarcameracalibration_2022}
J.~Beltrán, C.~Guindel, A.~de~la Escalera, and F.~García, ``{Automatic
  Extrinsic Calibration Method for LiDAR and Camera Sensor Setups},''
  \emph{{IEEE} Transactions on Intelligent Transportation Systems}, 2022.

\bibitem{peng_ppliteseg_2022}
J.~Peng, Y.~Liu, S.~Tang, Y.~Hao, L.~Chu, G.~Chen, Z.~Wu, Z.~Chen, Z.~Yu,
  Y.~Du, Q.~Dang, B.~Lai, Q.~Liu, X.~Hu, D.~Yu, and Y.~Ma, ``{PP-LiteSeg: A
  Superior Real-Time Semantic Segmentation Model},'' 2022.

\bibitem{pan_ddrnet_2022}
H.~Pan, Y.~Hong, W.~Sun, and Y.~Jia, ``{Deep Dual-Resolution Networks for
  Real-Time and Accurate Semantic Segmentation of Traffic Scenes},''
  \emph{{IEEE} Transactions on Intelligent Transportation Systems}, 2022.

\bibitem{cheng2021mask2former}
B.~Cheng, I.~Misra, A.~G. Schwing, A.~Kirillov, and R.~Girdhar,
  ``Masked-attention mask transformer for universal image segmentation,'' 2022.

\bibitem{hou_pointvoxelknowledgedistill_2022}
Y.~Hou, X.~Zhu, Y.~Ma, C.~C. Loy, and Y.~Li, ``{Point-to-Voxel Knowledge
  Distillation for LiDAR Semantic Segmentation},'' in \emph{{Proceedings of
  IEEE Conference on Computer Vision and Pattern Recognition (CVPR)}}, 2022.

\bibitem{zhu_cylinder3d_2021}
X.~Zhu, H.~Zhou, T.~Wang, F.~Hong, Y.~Ma, W.~Li, H.~Li, and D.~Lin,
  ``{Cylindrical and Asymmetrical 3D Convolution Networks for LiDAR
  Segmentation},'' in \emph{{Proceedings of IEEE Conference on Computer Vision
  and Pattern Recognition (CVPR)}}, 2021.

\bibitem{tang_spvnas_2020}
H.~Tang, Z.~Liu, S.~Zhao, Y.~Lin, J.~Lin, H.~Wang, and S.~Han, ``{Searching
  Efficient 3D Architectures with Sparse Point-Voxel Convolution},'' in
  \emph{{Proceedings of European Conference on Computer Vision (ECCV)}}, 2020.

\bibitem{yu_bisenet_2018}
C.~Yu, J.~Wang, C.~Peng, C.~Gao, G.~Yu, and N.~Sang, ``Bisenet: Bilateral
  segmentation network for real-time semantic segmentation,'' in
  \emph{{Proceedings of European Conference on Computer Vision (ECCV)}}, 2018.

\end{thebibliography}
